\documentclass{article}

\usepackage[final]{neurips_2022}
\usepackage{hyperref}       
\hypersetup{
    colorlinks = true,
    linkcolor = magenta,
    anchorcolor = blue,
    citecolor = cyan,
    filecolor = magenta,
    urlcolor = blue,
    pdftitle = {Measures of Information Reflect Memorization Patterns},
}

\usepackage[utf8]{inputenc} 
\usepackage[T1]{fontenc}    
\usepackage{url}            
\usepackage{booktabs}       
\usepackage{amsfonts}       
\usepackage{nicefrac}       
\usepackage{microtype}      
\usepackage{wrapfig}
\usepackage{algpseudocode}
\usepackage[table,xcdraw,dvipsnames]{xcolor}
\usepackage[outline]{contour}
\usepackage{multirow}
\makeatother
\usepackage{algorithm}
\usepackage{microtype}
\usepackage{graphicx}
\usepackage{booktabs} 
\usepackage{amsmath}
\usepackage{graphicx}
\usepackage{caption}
\usepackage{bbding}
\usepackage{multicol}
\usepackage[labelformat=empty, skip=0pt]{subcaption}
\usepackage{amssymb}
\usepackage{bbm}
\usepackage{hyperref} 
\usepackage{cleveref}
\usepackage{mathtools}
\usepackage[textsize=small]{todonotes}
\usepackage{enumitem}
\usepackage{manfnt}
\usepackage{wasysym}
\usepackage{color,soul}

\algnewcommand{\LineComment}[1]{\State \(\triangleright\) #1}
\newcommand*\emptysubfigure[1]{\begin{subfigure}[]{0pt}\caption{}\label{#1}\end{subfigure}}

\crefformat{section}{\S#2#1#3} 
\crefformat{subsection}{\S#2#1#3}
\crefformat{subsubsection}{\S#2#1#3}

\definecolor{beaublue}{rgb}{0.74, 0.83, 0.9}
\definecolor{lightapricot}{rgb}{0.99, 0.84, 0.69}

\newcommand{\memi}{example-level}
\newcommand{\memii}{heuristic}
\newcommand{\divi}{intra-neuron}
\newcommand{\divii}{inter-neuron}

\newcommand{\memiiSynthSetIN}{Colored MNIST}
\newcommand{\memiiSynthSetIIN}{Sentiment Adjectives}
\newcommand{\memiiSynthSetI}{\texttt{\memiiSynthSetIN}}
\newcommand{\memiiSynthSetII}{\texttt{\memiiSynthSetIIN}}

\newcommand{\memiiNatSetI}{\texttt{Bias-in-Bios}}
\newcommand{\memiiNatSetII}{\texttt{NICO$^{++}$}}
\newcommand{\memiSynthSetI}{\texttt{MNIST}}
\newcommand{\memiSynthSetII}{\texttt{IMDb}}

\makeatletter
\renewcommand*{\@fnsymbol}[1]{\ensuremath{\ifcase#1\or \jupiter\or \saturn\or \uranus\or \text{\manstar}\or \text{\manrotatedquadrifolium}\or *\or \dagger\or \ddagger\or
   \mathsection\or \mathparagraph\or \|\or **\or \dagger\dagger
   \or \ddagger\ddagger \else\@ctrerr\fi}}
\makeatother

\title{Measures of Information Reflect \\ Memorization Patterns}

\author{%
    Rachit Bansal\thanks{Work done during a visit at the Technion, Israel. The author is now at Google Research India.} \\
    Delhi Technological University \\
    \texttt{racbansa@gmail.com} \\
    \And
    Danish Pruthi\thanks{Work done while at Carnegie Mellon University, prior to joining Amazon.} \\
    Amazon Web Services \\
    \texttt{danish@hey.com} \\
    \And
    Yonatan Belinkov\thanks{Supported by the Viterbi Fellowship in the Center for Computer Engineering at the Technion.} \\
    Technion -- Israel Institute of Technology \\
    \texttt{belinkov@technion.ac.il} \\
}

\begin{document}

\maketitle
\linepenalty=1000

\setcounter{footnote}{0}

\begin{abstract}


Neural networks 
are known to
exploit spurious artifacts (or shortcuts)
that co-occur with a target label, exhibiting 
\emph{\memii} \emph{memorization}.
On the other hand, 
networks have been shown to memorize
training 
examples, resulting in \emph{\memi} \emph{memorization}.
These kinds of memorization impede generalization of networks
beyond their training distributions.
Detecting such memorization could be challenging, often 
requiring 
researchers to curate tailored test sets. 
In this work, we hypothesize---and subsequently show---%
that the diversity in the activation 
patterns of different neurons
is reflective of model generalization and memorization.
We quantify the diversity in the neural activations
through information-theoretic measures and
find support for our hypothesis in experiments spanning 
several natural language and vision tasks.
Importantly, we discover that information organization 
points to the two forms of memorization, 
even for neural activations computed on 
unlabeled in-distribution examples.
Lastly, we demonstrate the utility of our findings 
for the problem of model selection.
The associated code and other resources
for this work are available at~\href{https://information-measures.cs.technion.ac.il}{\url{https://rachitbansal.github.io/information-measures}}.

\end{abstract}
\section{Introduction}
\label{sec:intro}


Current day deep learning networks are limited in their 
ability to generalize across different domains and settings.
Prior studies
found that these 
networks 
rely 
on spurious artifacts 
that are correlated with a target label~\citep[][inter alia]{scholkopf2012causal, lapuschkin2019challenge, geirhos2019cnns, geirhos2020shortcut}.
We refer to learning of such artifacts (also known as heuristics or shortcuts)
as \textit{\memii\ memorization}.
Further, 
neural networks can 
also memorize
individual training examples and their labels; 
for instance, when a subset of the examples are incorrectly labeled
\citep{zhang_understanding_2017, arpit2018memorization, tanzer2021bertmem}.
We refer to this behavior as \emph{\memi\ memorization}.
A large body of past work has established
that these facets of memorization pose a
threat to generalization, especially in
out-of-distribution (OOD) scenarios 
where the memorized input features and
corresponding target mappings do not hold 
\citep{david2010ood, wang2021ood, hendrycks2021ood, shen2021ood}.
To simulate such OOD distributions, however,
researchers are required to laboriously 
collect specialized and labeled datasets 
to measure the extent of suspected fallacies in models.
While these sets make it possible to assess
model behavior over a chosen set of features,
the larger remaining features remain hard to
identify and study.
Moreover, these sets are truly extrinsic in nature,
necessitating the use of performance measures,
which in turn lack interpretability
and are not indicative of internal workings
that manifest certain model behaviors.
%
These considerations motivate 
evaluation strategies
that are intrinsic to a network and
indicate model generalization
while not posing practical bottlenecks
in terms of specialized labeled sets.
Here, we study information organization
as one such potential strategy.

%

\begin{figure*}[t]
    \centering
    \includegraphics[width=\textwidth]{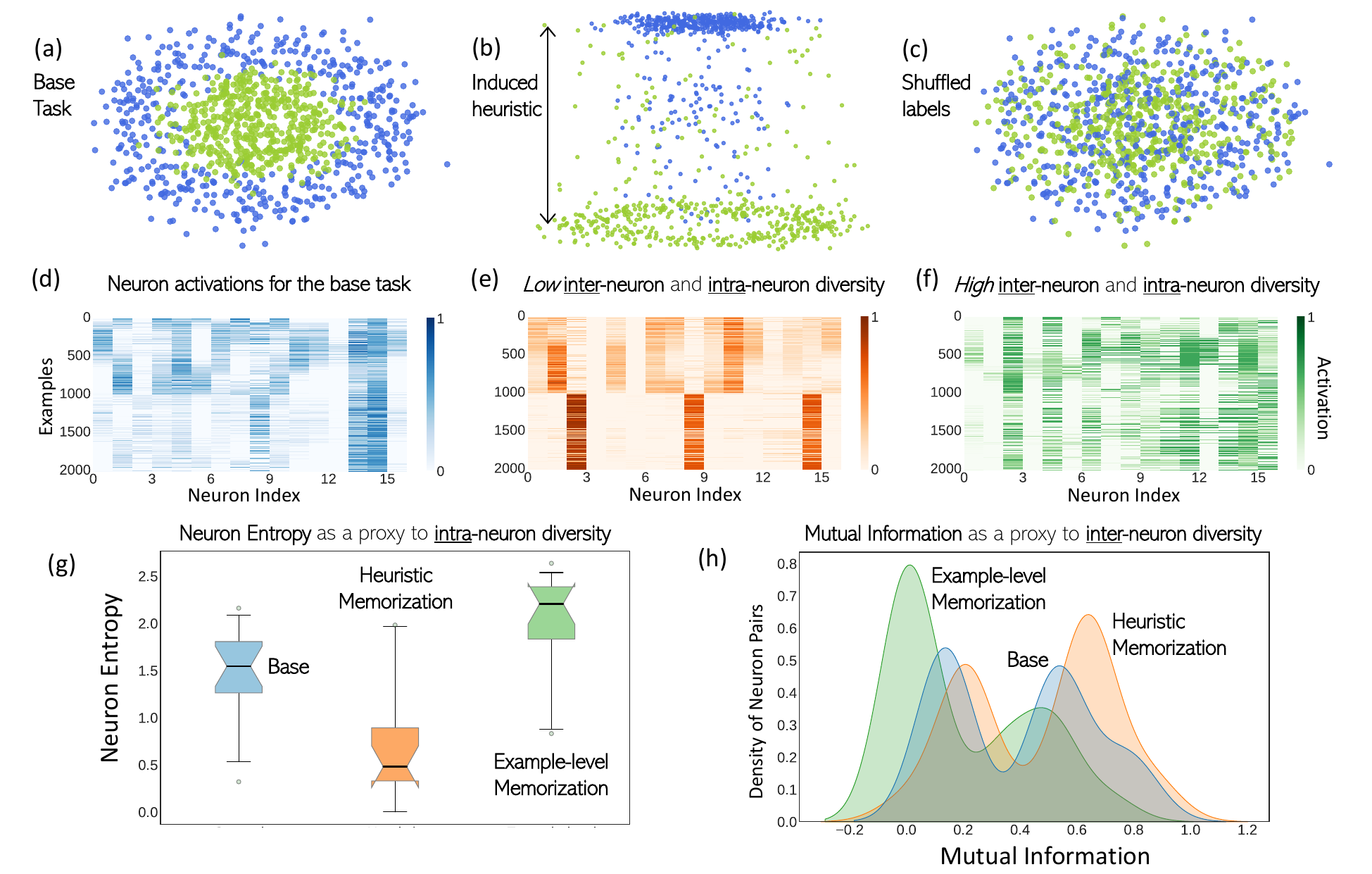}
    \emptysubfigure{sfig:circles_base} %
    \emptysubfigure{sfig:circles_funcmem} %
    \emptysubfigure{sfig:circles_exmem} %
    \emptysubfigure{sfig:circles_basediv} %
    \emptysubfigure{sfig:circles_lowdiv} %
    \emptysubfigure{sfig:circles_highdiv} %
    \emptysubfigure{sfig:circles_entropy} %
    \emptysubfigure{sfig:circles_mi} %
    \caption{
    (\textbf{a}) A toy setup of separating concentric circles;
    (\textbf{b}) An additional feature spuriously simplifies the task,
    inciting \emph{\memii\ memorization};
    (\textbf{c}) Shuffled target labels induce \emph{\memi\ memorization};
    (\textbf{d}) Neuron activations for a two-layered feed-forward network
    trained for the base task in (a);
    (\textbf{e}) Activation patterns for the network reflect 
    low \emph{\divi} and \emph{\divii} diversity when trained on (b);
    (\textbf{f}) High \emph{\divi} and \emph{\divii} diversity is seen 
    when the network is trained on (c);
    (\textbf{g}) \emph{Entropy} acts as a proxy to \divi\ diversity;
    (\textbf{h}) \emph{Mutual Information} acts as a proxy to \divii\ diversity.
    Distinguishable patterns for the three networks are seen in (g) and (h).
    \label{fig:circles}
    }
\end{figure*}

In this work, we posit that organization of information across
internal activations of a network
could be indicative of memorization.
Consider a sample task of separating concentric circles,
illustrated in Figure~\ref{sfig:circles_base}. 
A two-layered feed-forward network
can learn the 
circular decision boundary for this task.
However, if the nature of this learning task is changed, 
the network may resort to memorization.
When a spurious feature is introduced in this dataset
such that its value ($+$/$-$) correlates to the label ($0$/$1$)
(Figure~\ref{sfig:circles_funcmem}), 
the network \emph{memorizes} the feature-to-label mapping,
reflected in a 
uniform activation pattern across neurons
(Figure~\ref{sfig:circles_lowdiv}).
In contrast, when labels for the original set are shuffled 
(Figure~\ref{sfig:circles_exmem}),
the same network memorizes 
individual examples during training and 
shows a high amount of diversity in its
activation patterns (Figure~\ref{sfig:circles_highdiv}).
This example demonstrates how memorizing behavior 
is observed through diversity in neuron activations.

%

We formalize the notion of \emph{diversity} 
across neuron activations through two measures:
(i) \emph{\divi} diversity: the variation of activations
for a neuron across a set of examples, and 
(ii) \emph{\divii} diversity: the dissimilarity between pairwise
neuron activations on the same set of examples.
We hypothesize that
the nature of these quantities
for two networks
could point to underlying differences in 
their generalizing behavior.
In order to quantify \divi\ and \divii\ diversity,
we adopt the information-theoretic measures of 
\emph{entropy} and \emph{mutual information} (MI), 
 respectively.

%

Throughout this work, 
we investigate if diversity across neural activations~(\cref{sec:information})
reflects model generalizability.
We compare networks with varying levels of
\memii\ (\cref{sec:spurious_corr}) or 
\memi\ (\cref{sec:label_shuffling}) memorization
across a variety of settings:
 synthetic setups based on the 
\texttt{IMDb}~\citep{maas2011imdb} and 
\texttt{MNIST}~\citep{lecun1998mnist} datasets 
for both memorization types, as well as 
naturally occurring scenarios of
gender bias on  
\texttt{Bias-in-Bios}~\citep{de-arteaga2019biasinbios} 
and OOD image classification on \texttt{NICO}~\citep{zhang2022nico}.
We find that the information measures
consistently capture differences
among networks with varying degrees of
memorization:
Low entropy and high MI are characteristic of 
networks that show \memii\ memorization,
while high entropy and low MI are indicative of
\memi\ memorization.
Lastly, we evaluate these measures 
from the viewpoint of model selection
and note strong correlations 
to rankings from domain-specific evaluation 
metrics (\cref{sec:model_selection}).
\section{Methods}
\label{sec:information}

As per the data processing inequality~\citep{beaudry2012dpi}, 
a part of the neural network (referred to as the \textit{encoder})
compresses the most relevant information of a given input $X$,
into a representation $H$.
This compressed information is processed 
by a \textit{classification head} (or, a \textit{decoder})
to produce an output $Y$ corresponding to the given input.
We hypothesize that the organization of information across 
neurons of the encoder
is indicative of model generalization.
We study two complementary properties that
capture this information organization
for a given network:
\begin{enumerate}[label=(\roman*),topsep=1pt,itemsep=2pt,parsep=2pt]
    \item \textbf{Intra-neuron diversity}: How do activations of a given neuron vary across different input examples. 
    We measure the \emph{entropy} of neural activations (across examples) as a proxy. 
    \item \textbf{Inter-neuron diversity}: How unique is the activation of a neuron compared to other neurons. 
    We quantify this via \emph{mutual information} between activations of pairwise neurons. 
\end{enumerate}

Below, we discuss the information measures formally.

\subsection{Information Measures}
\label{subsec:inf_meas}
For any given encoder
(consisting of $N$ neurons)
that maps 
the input to a dense 
hidden representation, 
we denote 
the activation of the 
$i^{\text{th}}$ neuron 
as 
a random variable, $A_{i} \in \{a_{i}^{1}, \dots, a_{i}^{S}\}$,
where each measurement is an activation over an example from a set of size $S$.
The probability over this continuous activation space is computed
by binning it into discrete ranges~\citep{darbellay1999info}, and
we denote each discretized activation value as $\hat{a}$.
Importantly, the set of examples
on which the activations are computed
come from a distribution
that is similar to the underlying
training set itself.

\paragraph{Entropy}
We measure the Shannon entropy for each neuron in the concerned network,
as a proxy of \divi\ diversity.
Following the definition of Shannon entropy, 
this is given as:
\begin{align}
    & H(A_{i}) = \mathop{\mathbb{E}}_{\hat{a}_i^{s} \in A_{i}}[h(\hat{a}_i^{s})]
    = \sum_{j=1}^{N_{\text{bins}}}p(\hat{a}_i^{j})\log(\frac{1}{p(\hat{a}_i^{j})})
    \label{eq:entropy}
\end{align}


\paragraph{Mutual Information}
We compute the mutual information (MI) 
between underlying neurons
as a proxy to \divii\ diversity.
Specifically, we compute the MI between 
all neuron pairs in the network.%
\footnote{
In principle, we would 
compute MI across neuron sets; 
we approximate
this through individual neuron pairs.
}
Thus, the set of MI values $I(A_{i})$ for a particular neuron $A_{i}$, is given as:

\begin{align}
    & I(A_{i}) = \{I(A_{i}; A_1), \dotsc, I(A_{i}; A_{N})\} \label{eq:mi}
\end{align}
where, 
$I(X; Y)$ depicts the MI between variables $X$ and $Y$.
Unless stated otherwise, this $I(A_{i})$ is computed 
$\forall~i \in \{1, \dotsc, N\}$, 
resulting into a square matrix of size $(N \times N)$.


This process of computing the information measures
for a network on a given set of examples is summarized in
Algorithm \ref{alg:info_measure}. 
Further details on the computation 
are given in \cref{sec:computing}.

\begin{algorithm}[!ht]
    \caption{
    Computation of information measures.
    Algorithmic procedures 
    \textsc{Entropy} and \textsc{MI}
    are specified by algorithms \ref{alg:entropy_computation}
    and \ref{alg:mi_computation} in
    \cref{sec:computing}.
    \label{alg:info_measure}
    }
    \begin{algorithmic}[1]
        \State $A_1, \dots, A_{N} \gets \{f(x_i)\}_{i=1}^{S}$
        \Comment{Computing activations for all neurons}
        \State $H \gets \{\}$; $I \gets \{\}$
        \Comment{Initiating computations for Entropy and MI}
        \For{$i \in \{1, \dots, N\}$}
        \Comment{Iterating over the set of neurons}
            \State $I_{i} \gets \{\}$
            \Comment{Initiating MI for a particular neuron}
            \State $H_{i} \gets \textsc{Entropy}(A_i)$
            \Comment{Following Equation~\ref{eq:entropy} and Algorithm~\ref{alg:entropy_computation}}
            \For{$j \in \{1, \dots, N\}$}
            \Comment{Inner loop over the set of neurons}
                \State $I_{i} \gets I_{i} \bigoplus \textsc{MI}(A_i, A_j)$
                \Comment{Following Equation~\ref{eq:kraskov} and Algorithm~\ref{alg:mi_computation}}
            \EndFor
            \State $H \gets H \bigoplus H_{i}$
            \State $I \gets I \bigoplus I_{i}$
            \Comment{Following Equation~\ref{eq:mi}}
        \EndFor
    \end{algorithmic}
\end{algorithm}

\subsection{Toy Setup: Concentric Circles}
\label{subsec:concentric_circles}



Here, we briefly discuss the information-theoretic metrics for 
the example of concentric circles from the introduction (Figure \ref{fig:circles}). 
To recap, we consider a setup to compare networks
showing the two forms of memorization and observe
discernible differences in their activation
patterns: \memii\ memorization corresponds to low \divi\ and \divii\ diversity,
while \memi\ memorization corresponds to high diversity 
(Figures \ref{sfig:circles_lowdiv} and \ref{sfig:circles_highdiv}).
We expect that this difference in diversity would be captured
through the above defined information measures. 

Figure \ref{sfig:circles_entropy} presents the distribution of entropy values 
for each of the three networks
with varying generalization behaviors.
Throughout this work, we visualize this distribution of entropy 
using similar box-plots, where a black marker within the boxes
depicts the median of the distribution and a
notch neighboring this marker depicts
the $95\%$ confidence interval around the median.
We observe that 
entropy for the network exhibiting
\memii\ memorization is distributed around a lower point than the others,
whereas entropy for the network with \memi\ memorization is higher.

Furthermore, Figure \ref{sfig:circles_mi} shows the distribution of MI
for the three networks. 
To interpret the distribution of MI (an $N\times N$ square matrix),
we fit a Gaussian mixture model
over all values 
and 
visualize it through a density plot,
where the density (\emph{y-axis}) at each point corresponds to the number
of neurons pairs that exhibit that MI value (\emph{x-axis}).
Larger peaks in these density plots 
suggest a large number of neurons pairs 
are concentrated in that region.
Interestingly, we see such peaks for the three networks
at distinct values of MI. For the network showing \memi\ memorization
(high \divii\ diversity),
most of the neuron pairs show low values of MI. In contrast,
\memii\ memorization (low \divii\ diversity) 
has high neuron pair density for higher MI values.\footnote{
This difference in neuron activation patterns for the two memorizing sets
could be caused by several factors, including functional complexity~\citep{lee2020complexity}:
Functions that encode individual data points (as in \memi\ memorization) need to be much more complex
than functions that learn shortcuts (\memii\ memorization).
We make a comparison with standard complexity
measures in \cref{subsec:complexity_measures}
and observe that our information measures correlate more
strongly with generalization performance---especially 
for \memii\ memorization.
}

Based on these findings, we formulate two hypotheses, summarized in Table~\ref{tb:relationships}: 

\begin{table}[ht]
\begin{minipage}{.53\textwidth}
\begin{enumerate}[leftmargin=*,itemsep=2pt,topsep=0pt,parsep=1pt,label=\textbf{H\arabic*}]
    \item \label{item:hyp_1} Networks exhibiting 
    \memii\ memorization would show
    low inter- and intra-neuron diversity,
    reflected through low entropy and high MI values.
    \item \label{item:hyp_2} Networks  exhibiting 
    \memi\ memorization would show
    high inter- and intra-neuron diversity,
    reflected through high entropy and low MI values. 
\end{enumerate}
\end{minipage} \hspace{3pt}
 \begin{minipage}{.47\textwidth}
\vspace{-15pt}
\centering
\caption{
Summarizing our hypotheses.
\label{tb:relationships}
}
\setlength\tabcolsep{3pt}
\begin{tabular}{l|cc}
\toprule
    \multicolumn{1}{c|}{Memorization} 
    & \multicolumn{2}{c}{Diversity} \\ \midrule
        & \multicolumn{1}{c}{\cellcolor[HTML]{F1F7FF}\begin{tabular}[c]{@{}c@{}}Intra-neuron\\($\propto$ Entropy)\end{tabular}}
        & \multicolumn{1}{c}{\cellcolor[HTML]{FFEBD3}\begin{tabular}[c]{@{}c@{}}Inter-neuron\\($\propto$ MI$^{-1}$)\end{tabular}}  \\
      \cline{2-3}
            Heuristic
                & \cellcolor[HTML]{F1F7FF}\contour{black}{$\downarrow$}
                & \cellcolor[HTML]{FFEBD3}\contour{black}{$\downarrow$} \\
            Example-level
                & \cellcolor[HTML]{F1F7FF}\contour{black}{$\uparrow$} 
                & \cellcolor[HTML]{FFEBD3}\contour{black}{$\uparrow$}
            \\ \bottomrule
\end{tabular}
\end{minipage}
\end{table}

\vspace{-5pt}
\section{Heuristic Memorization}
\label{sec:spurious_corr}
\vspace{-5pt}

Here, we study 
different networks with varying degrees of
\memii\ memorization,
and examine if the information
measures---aimed to capture neuron diversity---indicate the 
extent of memorization.

\subsection{Semi-synthetic Setups}
\label{subsec:semi_synthetic_memii}

We synthetically introduce spurious artifacts 
in the training examples 
such that they co-occur with target labels.
Networks trained on such a set are prone to 
memorizing these artifacts.
The same correlations with an artifact do not hold in the validation sets.
To obtain a set of networks 
with varying degrees of this \memii\ memorization, 
we consider a parameter $\alpha$ that controls
the fraction of the training examples 
for which
the spurious correlation holds true.
We consider the following setups:

\paragraph{\memiiSynthSetIN}
In this setting, the MNIST dataset~\citep{lecun1998mnist} 
is configured such that a network trained on this set 
simply learns to identify the color of images and 
not the digits themselves~\citep{arjovsky2019irm}.
Particularly, digits $0$--$4$ are 
grouped as one label while $5$--$9$ as the other, 
and images for these labels are colored green and red, respectively. 
For this setup, we train multi-layer perceptron (MLP) 
networks for varying values of $\alpha$, 
which corresponds to the fraction of training 
instances that abide to the color-to-label correlation.
The considered values of $\alpha$ and other details for this
setup are given in \cref{subsec:details_colored_mnist}.

\paragraph{\memiiSynthSetIIN}
In this setup, 
we sub-sample examples 
from the \texttt{IMDb} dataset~\citep{maas2011imdb} 
that contain at least one adjective from a list of 
positive and negative adjectives.
Then, examples that contain any of the positive adjectives 
(``good'', ``great'', etc.)
are marked with the positive label, 
whereas ones that contain any negative adjectives 
(``bad'', ``awful'', etc.)
are labeled as negative. 
We exclude examples that contain adjectives
from both lists. 
The motivation to use this setup is to introduce  
heuristics in the form of adjectives in the training set.
We fine-tune DistilBERT-base models~\citep{sanh2019distilbert} 
on this task for different values of $\alpha$
(fraction of examples that obey the heuristic).
The full set of adjectives considered and further details
are outlined in \cref{subsec:details_adjectives}.

\begin{figure*}[t]
    \centering
    \begin{subfigure}[t]{0.49\linewidth}
        \centering
        \includegraphics[width=\textwidth]{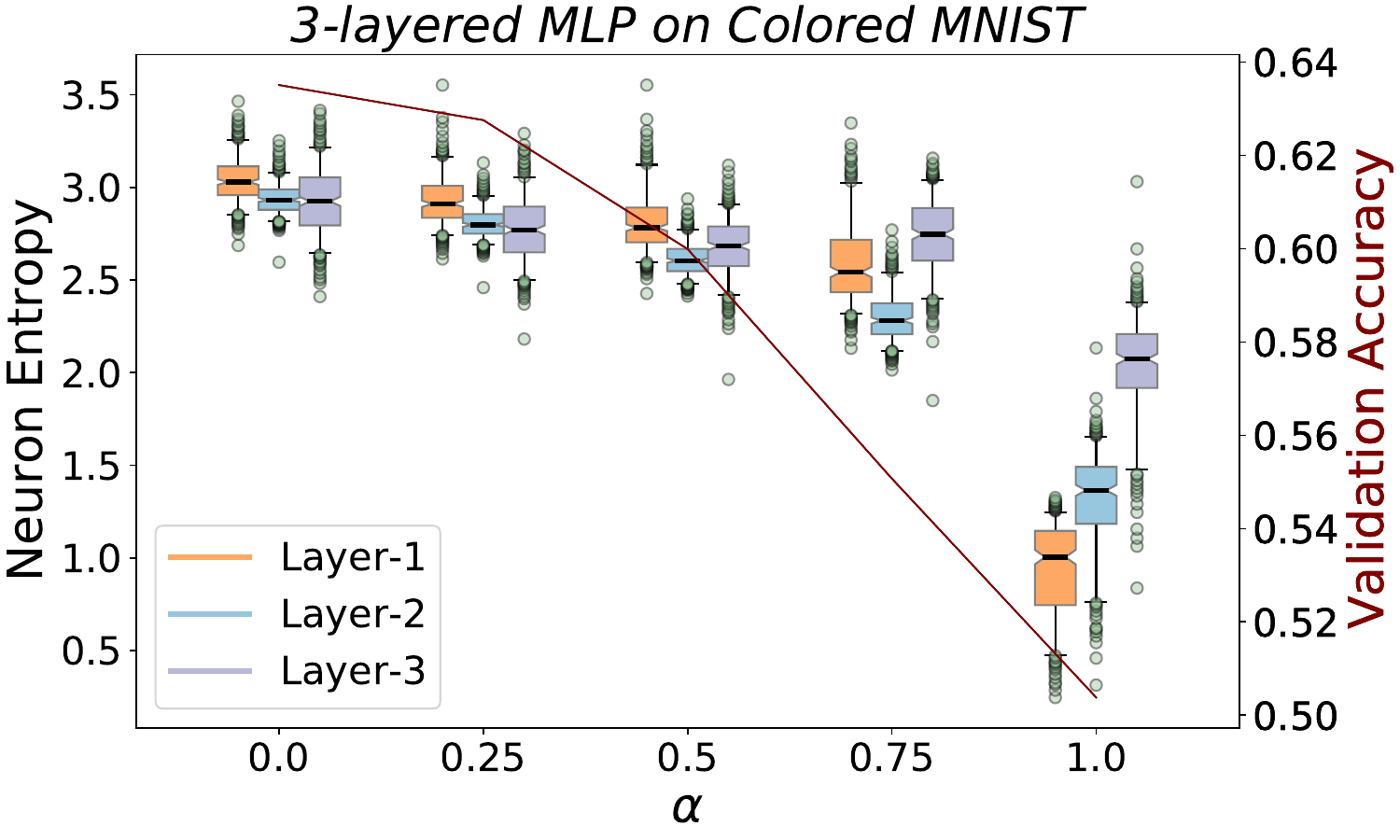}
        \label{sfig:colored_mnist_entropy}
    \end{subfigure}
    \hfill
    \begin{subfigure}[t]{0.49\linewidth}
        \centering
        \includegraphics[width=\textwidth]{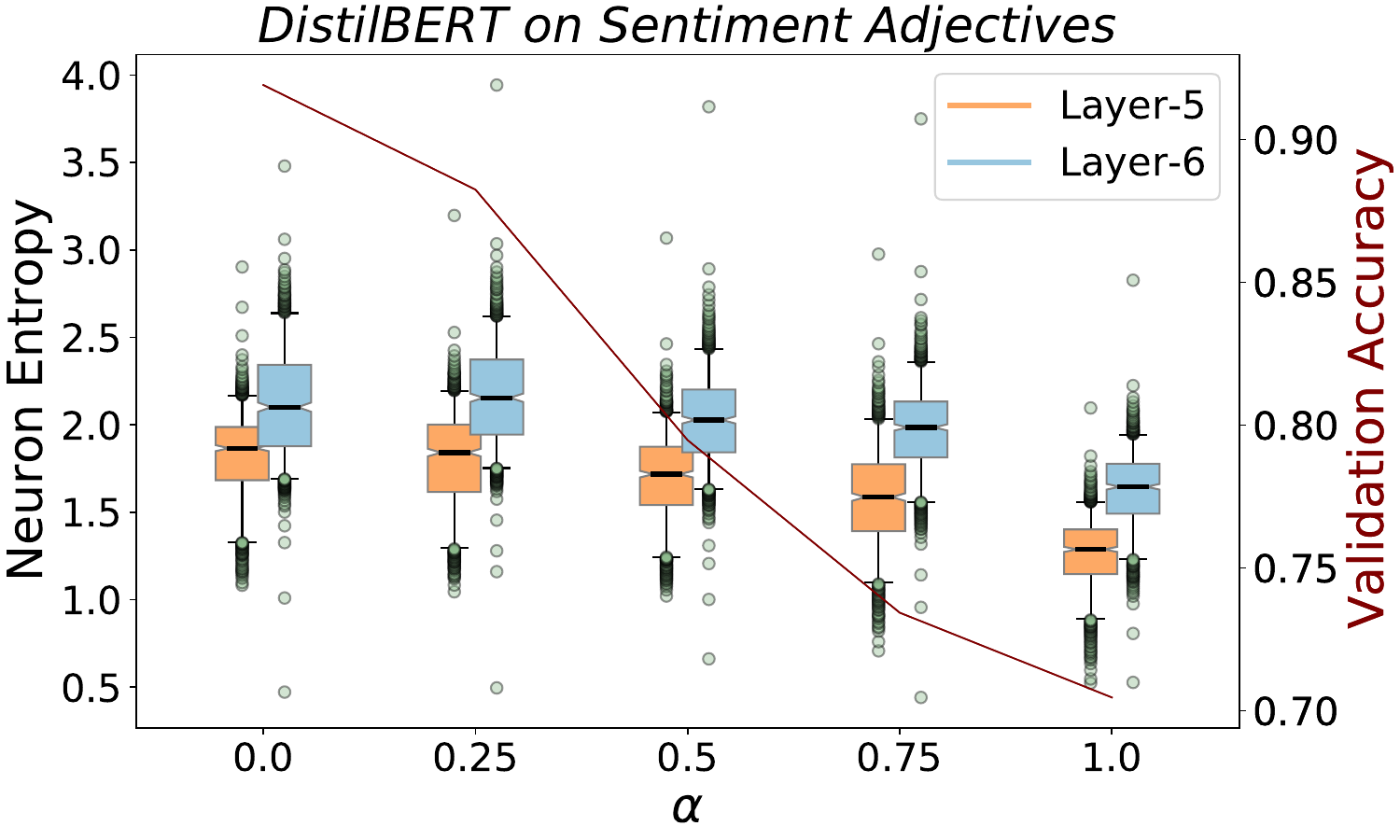}
        \label{sfig:imdb_adjectives_entropy}
    \end{subfigure}
    \caption{
    The relation between entropy of neural activations 
    and \memii\ memorization.
    For both the setups, 
    networks trained on higher $\alpha$
    show higher \memii\ memorization
    (as depicted by the dipping
    model accuracy {\color{Maroon} line}),
    accompanied with lower entropy values.
    \label{fig:semi_synthetic_spurious}
    }
\end{figure*}

\begin{figure*}[t]
    \centering
    \begin{subfigure}[t]{0.48\linewidth}
        \centering
        \includegraphics[width=\textwidth]{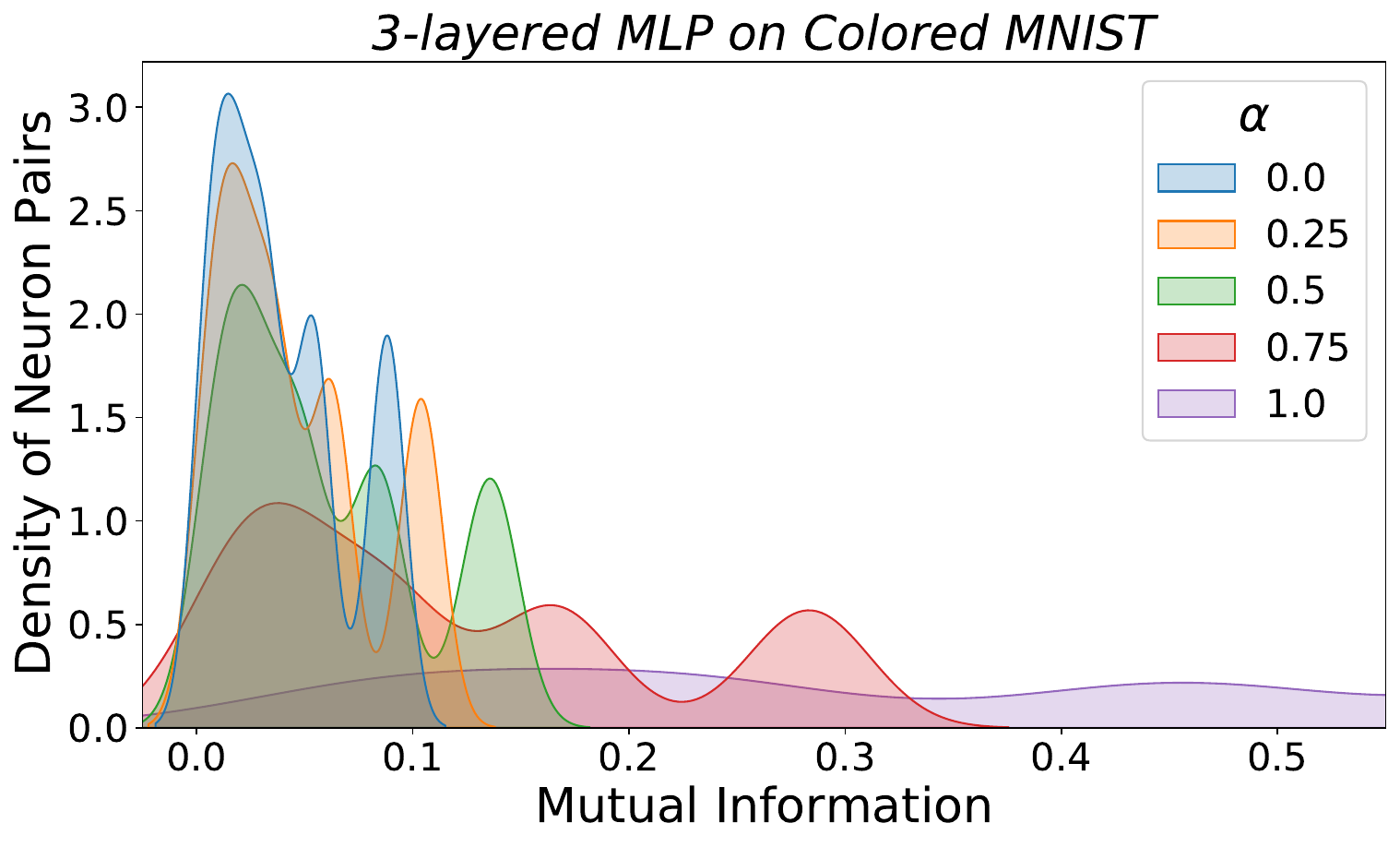}
        \label{sfig:colored_mnist_mi}
    \end{subfigure}
    \hfill
    \begin{subfigure}[t]{0.48\linewidth}
        \centering
        \includegraphics[width=\textwidth]{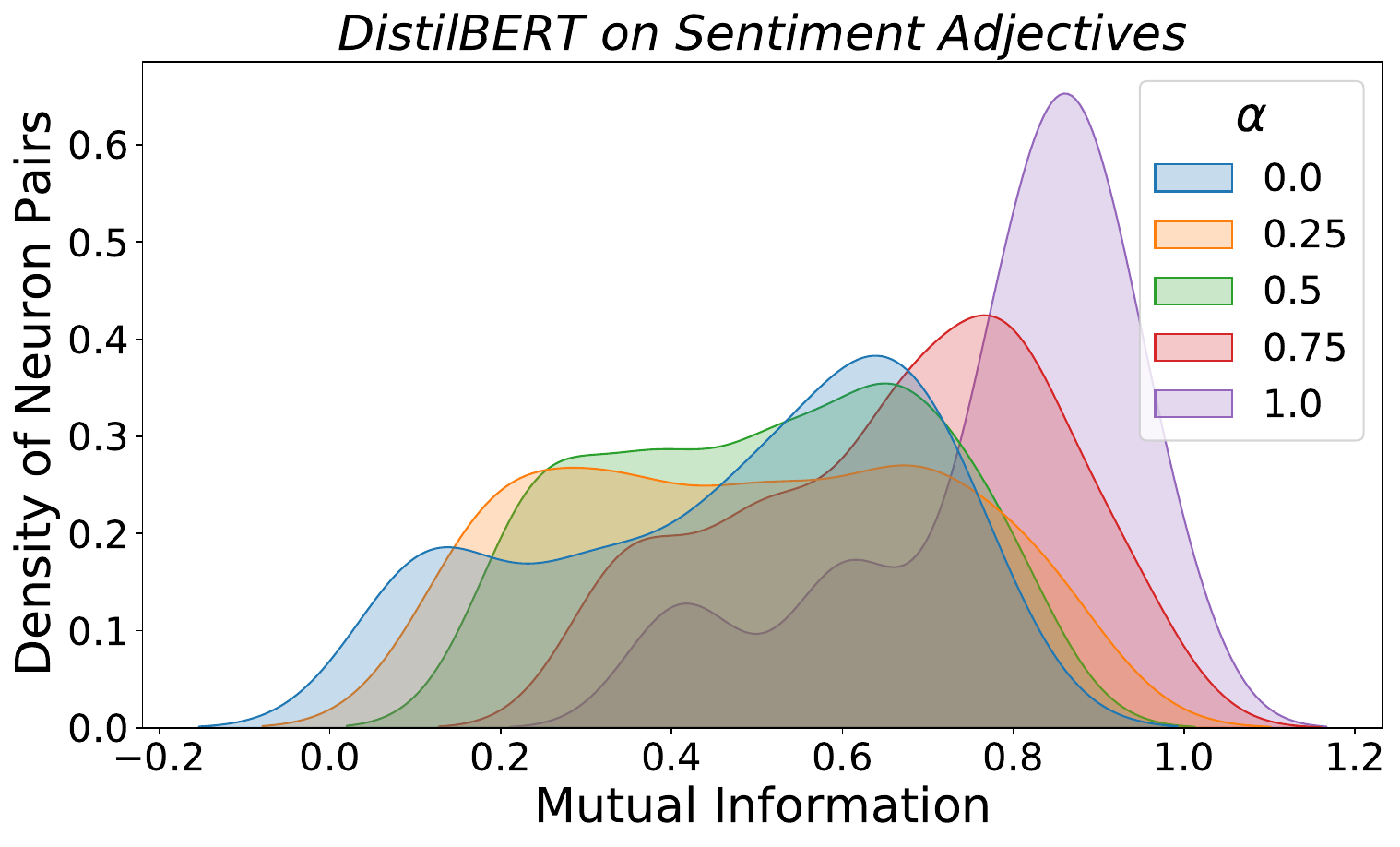}
        \label{sfig:imdb_adjectives_mi}
    \end{subfigure}
    \caption{
    Distribution of mutual information (MI) of pairs of neurons
    for networks with varying heuristic memorization.
    For both settings,
    networks trained on training sets
    with larger amounts of spurious correlations
    ($\uparrow \alpha$)
    exhibit higher mutual information across their neuron pairs.
    }
    \label{fig:semi_synthetic_spurious_mi}
\end{figure*}

\paragraph{Results:} Through these experiments, we first note that
\textbf{low entropy across neural activations 
indicates heuristic memorization in networks}.
This is evident from Figure~\ref{fig:semi_synthetic_spurious},
where we see that (1) as we increase $\alpha$ 
the validation performance decreases, indicating 
heuristic memorization (see the solid line in the plots); and 
(2) with an increase in this 
{\memii} memorization, 
we see lower entropy across neural activations. 
We show the entropy values of neural activations for 
the $3$ layers of an MLP
trained on
\memiiSynthSetI\ (left sub-plot) 
and 
for the last two layers of 
DistilBERT
on \memiiSynthSetII\ (right sub-plot).\footnote{
Considerable changes in entropy values are not seen
for initial DistilBERT layers, suggesting that 
spurious correlations are largely
captured by later layers. 
Detailed results covering other layers are given in 
\cref{subsec:semi_synthetic_memii_all_results}.
}
In both these two scenarios, we see a consistent drop in the 
entropy with increasing values of $\alpha$, with a particularly 
sharp decline when $\alpha=1.0$.

Furthermore, we observe that 
\textbf{networks with higher heuristic memorization 
exhibit higher mutual information} across pairs of neurons.
In Figure~\ref{fig:semi_synthetic_spurious_mi}, 
networks with higher memorization ($\uparrow \alpha$), 
have larger density of neurons in the 
high mutual information region.
While this trend is consistent across the two settings,
we see some qualitative differences:
The memorizing ($\alpha = 1.0$) MLP network on \memiiSynthSetI\ (left)
has a uniform distribution across the entire scale of MI values,
while DistilBERT on \memiiSynthSetII\ (right) largely has a 
high-density peak for an MI of $\sim 0.9$.

\subsection{Natural Setups}
\label{subsec:natural_memii}

Next, we investigate setups 
where spurious correlations 
are not synthetically induced, but
occur naturally in the datasets.
Below, we describe two such scenarios:

\begin{figure*}[t]
    \centering
    \begin{subfigure}[t]{0.45\linewidth}
        \centering
        \includegraphics[width=\textwidth]{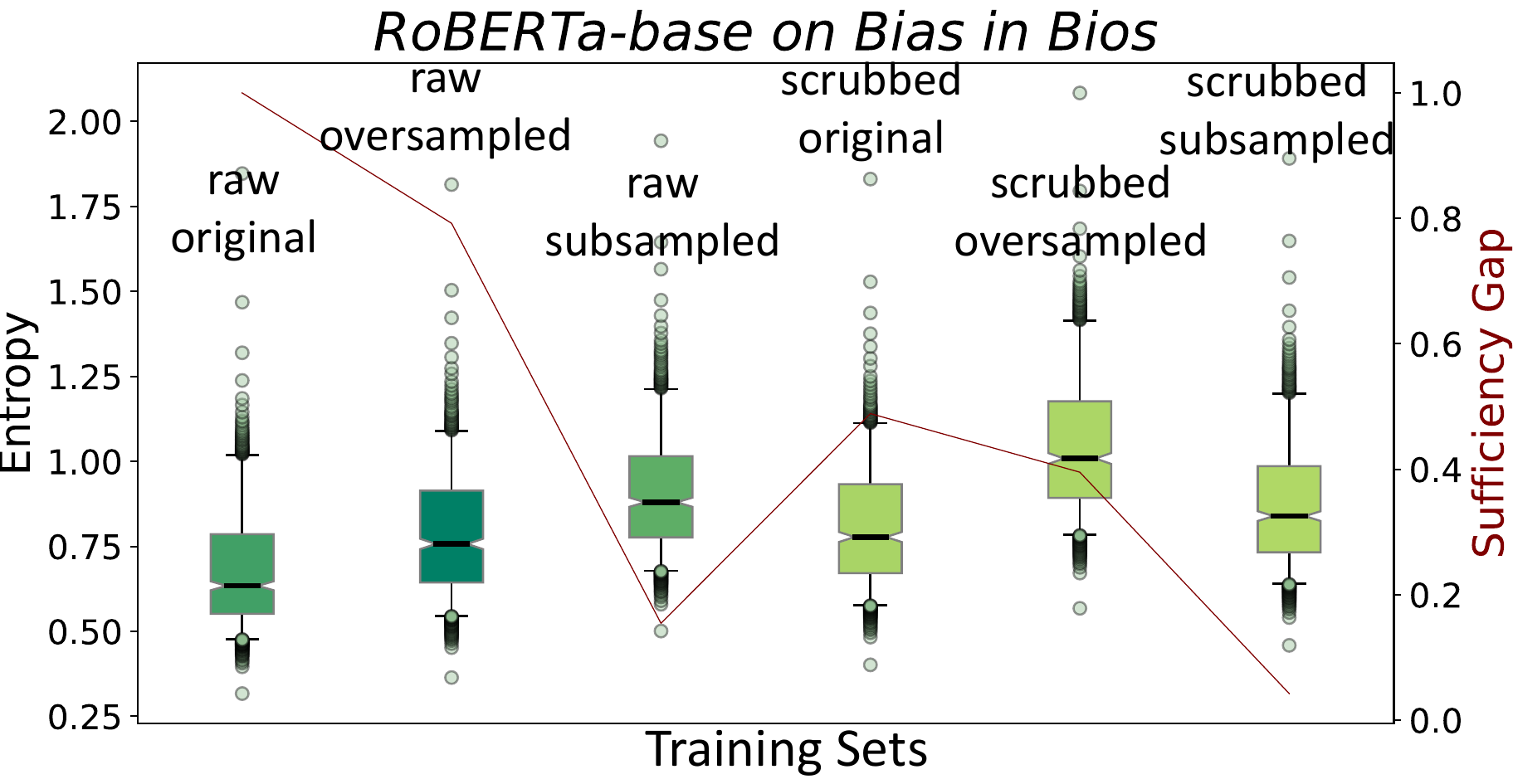}
    \end{subfigure}
    \hfill
    \begin{subfigure}[t]{0.45\linewidth}
        \centering
        \includegraphics[width=\textwidth]{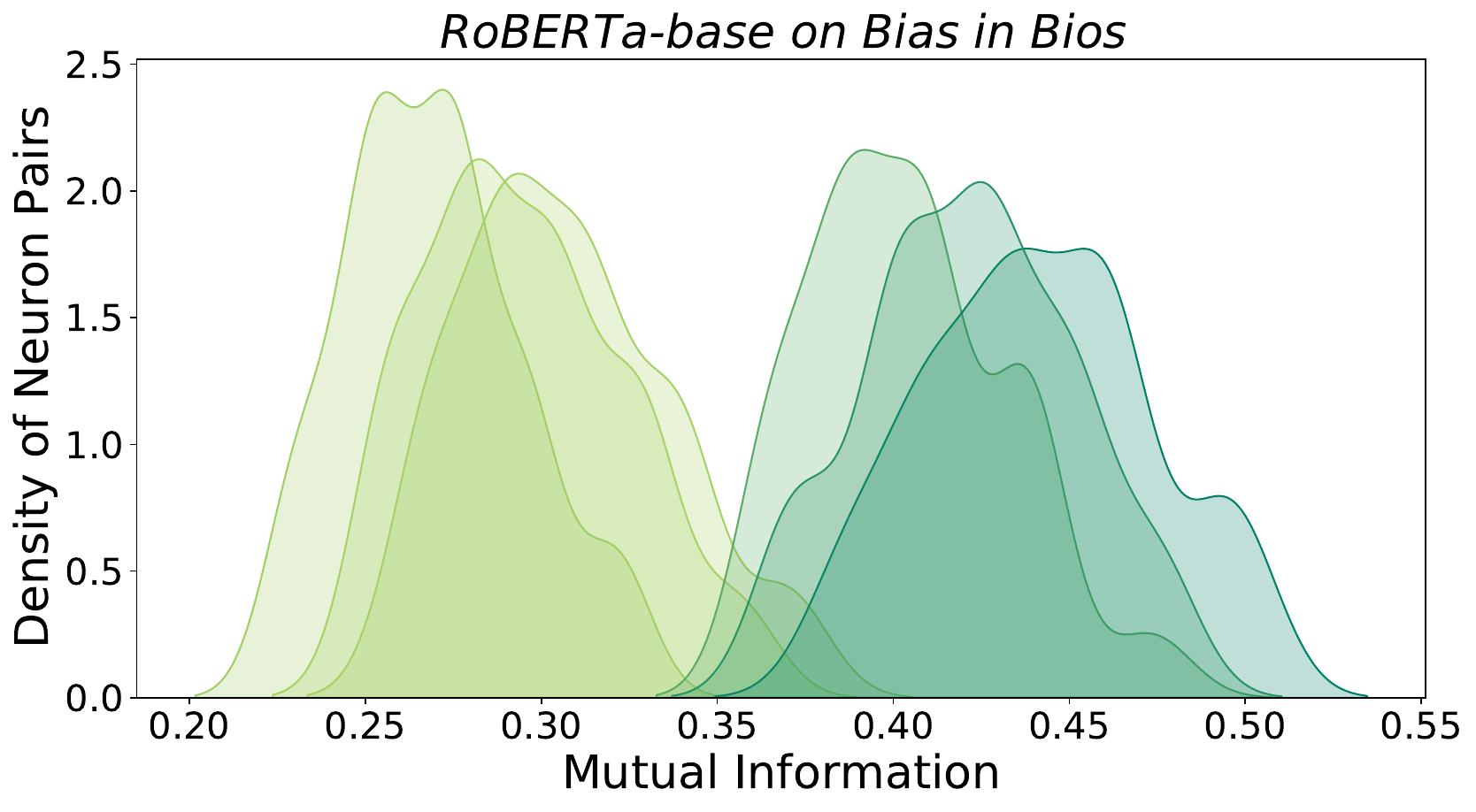}
    \end{subfigure}
    \hfill
    \begin{subfigure}[t]{0.075\linewidth}
        \centering
        \includegraphics[width=\textwidth]{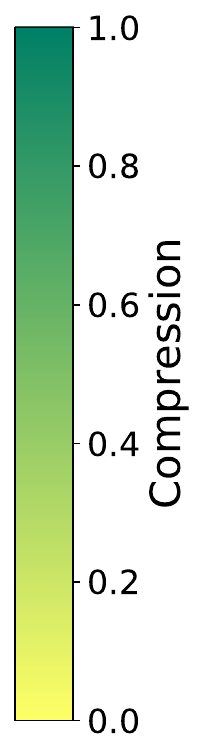}
    \end{subfigure}
    \caption{
    Distributions of entropy and MI across  final 
    layer activations of RoBERTa-base
    differentiate networks
    fine-tuned on original and de-biasing sets
    for \memiiNatSetI.
    Color of boxes and Gaussian plots corresponds to 
    \emph{extractability} of gender information in 
    model representations as estimated through 
    MDL probing~\citep{voita2020mdl}%
    ---lighter colors
    indicate lower extractability (less bias).
    }
    \label{fig:biasinbios}
\end{figure*}

\paragraph{Occupation Prediction}
\label{para:bias_in_bios}
We first study the task of predicting occupations from biographies  
on the \memiiNatSetI\ dataset~\citep{de-arteaga2019biasinbios}.
Given the skewed distribution of genders across occupations, 
models pick up
cues that reveal the biographee's gender.
For instance, most biographies 
corresponding to the ``professor'' occupation 
are of males.
Models trained on this dataset can
learn such spurious associations. 
To evaluate how much the trained 
networks encode gender, we measure
\textit{compression} values by
training a gender classifier 
on the internal representations of the network
and computing its minimum description length (MDL).
These compression values act as a 
proxy to the ease of extracting 
gender information
from representations~\citep{voita2020mdl,orgad2022debiasing}.

We consider a variety of training sets
by \emph{sub-sampling} and \emph{over-sampling} examples
for each profession in the dataset: This is done to balance 
the number of examples across each gender. 
We do this for both the original inputs 
in the dataset (\emph{raw}) and 
\emph{scrubbed} examples,
wherein gender-specific information (such as pronouns) is removed 
(similar to setups in~\citet{de-arteaga2019biasinbios}).
We perform our analysis on RoBERTa-base~\citep{liu2019roberta} fine-tuned 
for these training sets.%
\footnote{
We use trained checkpoints released by \citet{orgad2022debiasing}. 
More details are given in \cref{subsec:details_bias_bios}.
} 

\textbf{Results:} In Figure~\ref{fig:biasinbios},
we observe the distribution
of the two information measures for 
the last layer of networks
trained on the different training sets.%
\footnote{
The difference in compression values across training
sets is more prominent in higher layers,
yet the correlation between compression and MI
remains high throughout the network. 
We discuss this in \cref{subsec:bias_mi_all_layers}.
}
This variation is shown in conjunction with 
compression values across the network
using the MDL probe.
Following our initial hypothesis (Table~\ref{tb:relationships};~\ref{item:hyp_1}),
we expect that networks with higher representation of bias
will have lower entropy. 
Indeed, in Figure \ref{fig:biasinbios} (left),  
the network trained on the original training set 
(i.e., \texttt{raw original})
shows
the lowest entropy. 
This finding is in line with our hypothesis,
since the other networks are trained on either gender-balanced
or scrubbed sets.
However, we do not observe consistent trends
among networks trained on these de-biasing sets.
On the other hand, we do see clear patterns in MI 
that distinguish networks in line with their compression values 
(Figure \ref{fig:biasinbios}, right).
As we go from lower to higher values of MI
(left to right), the density plots get darker,
corresponding to higher compression values (higher bias).
A prominent distinction is seen between the \texttt{raw} and
\texttt{scrubbed} sets, which are separated on two sides of the plot.

\begin{figure*}[t]
    \centering
    \begin{subfigure}[t]{0.49\linewidth}
        \centering
        \includegraphics[width=\textwidth]{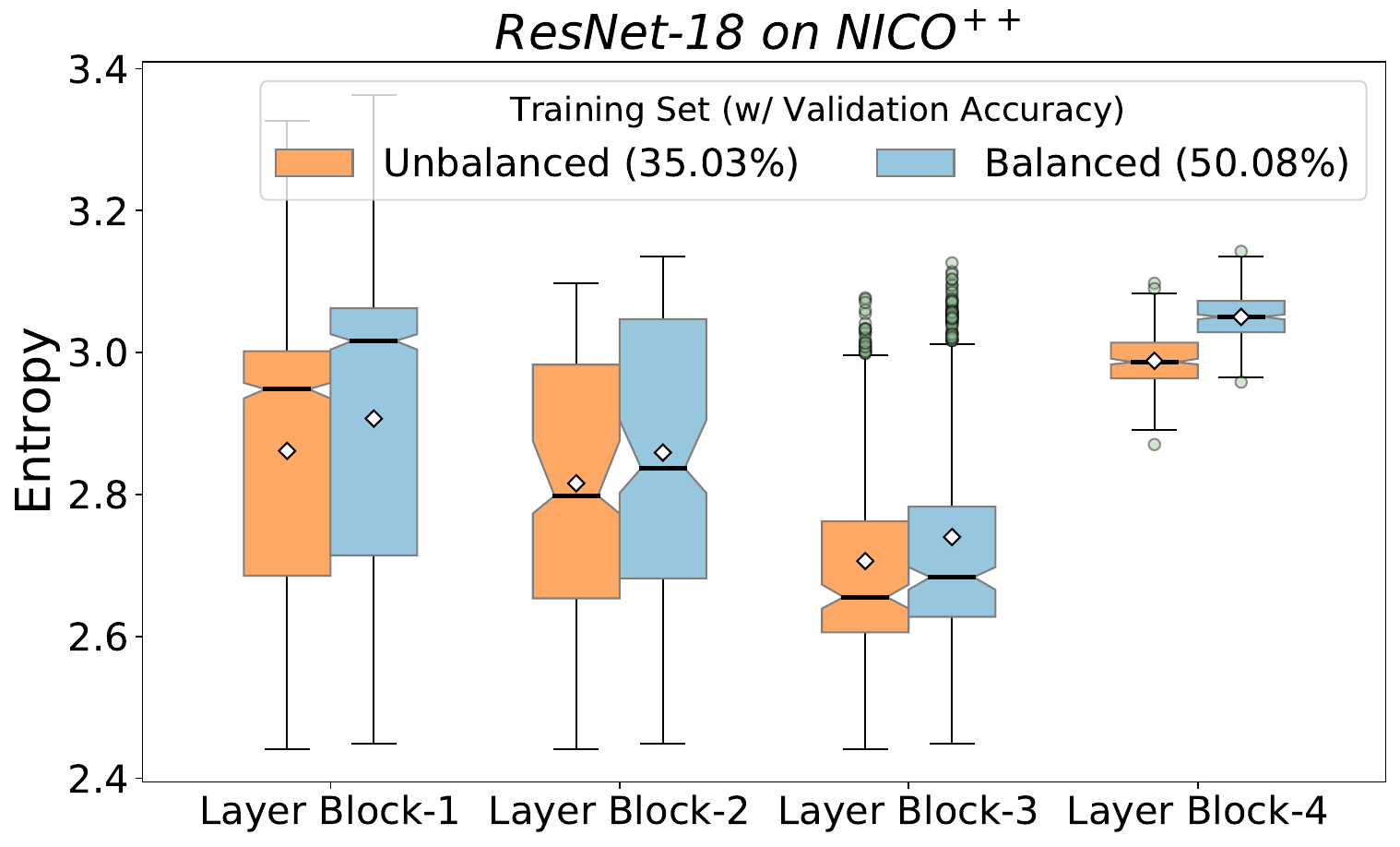}
    \end{subfigure}
    \hfill
    \begin{subfigure}[t]{0.49\linewidth}
        \centering
        \includegraphics[width=\textwidth]{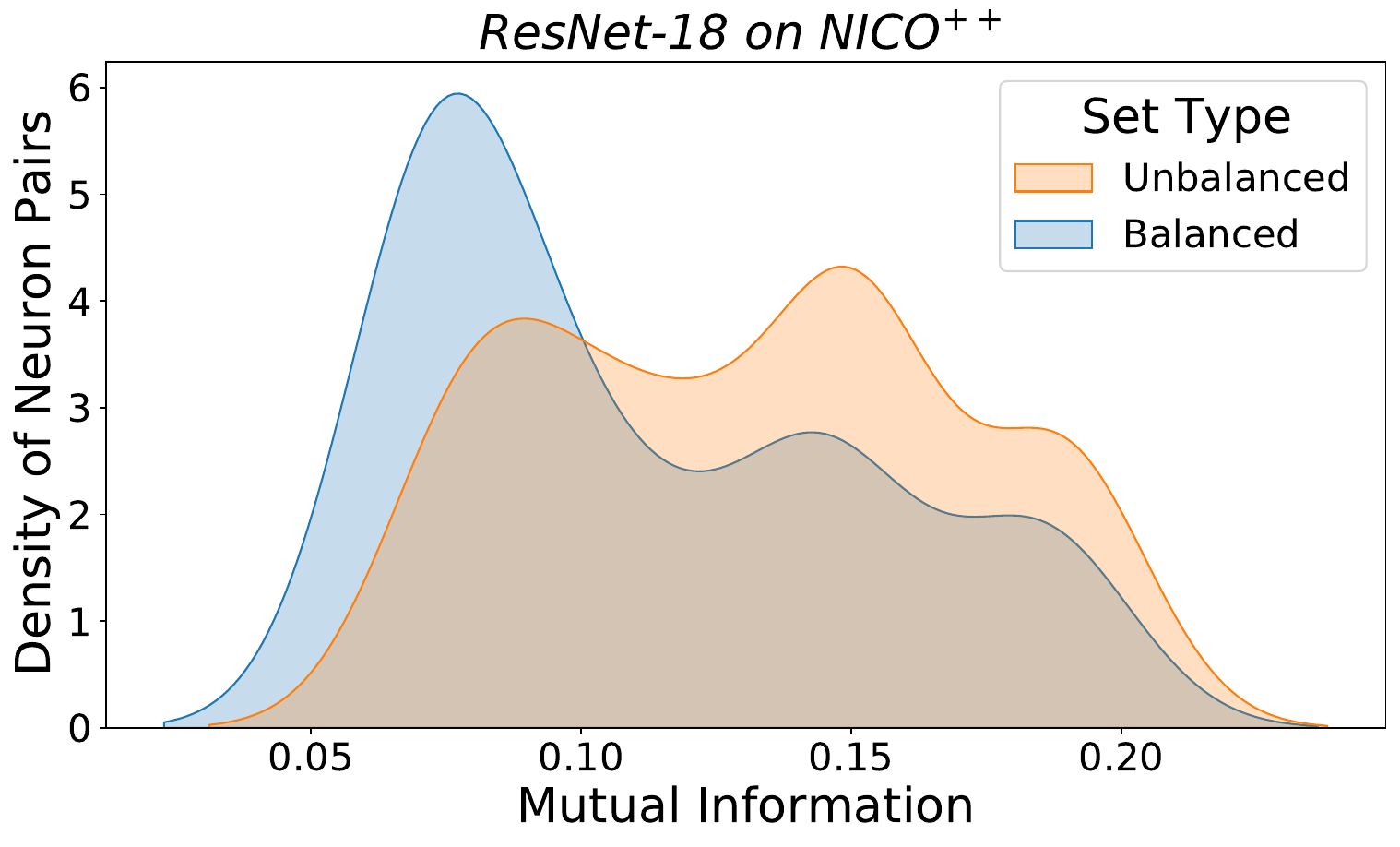}
    \end{subfigure}
    \caption{
    Entropy and MI for ResNet-18 on the \memiiNatSetII\ dataset.
    The two training sets---\texttt{balanced} and \texttt{unbalanced}---result
    into models that vary in their generalization to contextual features
    beyond on what they were trained on. This distinction is reflected in
    the information measurements.
    \label{fig:nico}
    }
\end{figure*}

\paragraph{Image Classification with Contexts}
\label{para:nico}
Next, we consider
a scenario from computer vision, 
where 
the task is to 
identify the presented object in a particular context.
We use a subset of 
the \memiiNatSetII dataset~\citep{zhang2022nico},
which consists of images of animals
in a variety of contexts.
For each animal class, there exist 
two types of contexts: 
\textit{individual}, those that are specific to only that animal
and are not present for all classes
(such as a \textit{roaring} bear),
and \textit{common}, contexts that exist
across all classes
(such as images taken in \textit{dark}).

For our analysis, we design two training 
sets---\texttt{unbalanced} and \texttt{balanced}---varying
in
the distribution of common contexts
across examples.
Each animal in the \texttt{unbalanced} set occurs
in a particular common context that is chosen for that animal.
In contrast, the \texttt{balanced} set contains 
images from all common contexts, for each animal.
Thus, a network trained on the \texttt{unbalanced} set
is likely to pick the context-to-animal mapping
(i.e., a case of \memii\ memorization). 

\textbf{Results:}
We train ResNet-18~\citep{he2015resnet} 
networks for the two sets
and evaluate them on the common \memiiNatSetII\
evaluation set, balanced across
all common contexts.
We consider the hidden representation
from each of the $4$ blocks of layers in the network
to compute the information measures 
reported in Figure \ref{fig:nico}.
From the left sub-figure, 
we observe that the entropy for networks
trained on the \texttt{balanced} set
is consistently greater than the \texttt{unbalanced}
set across all layer blocks.
Furthermore, we observe 
that distribution of MI (right)
across pairwise neurons 
also reflects the difference 
between the networks, 
corroborating our hypothesis.
Neuron pairs for the network that memorizes 
the correlation with image contexts (\texttt{unbalanced})
are more densely concentrated at higher MI values.
\section{Example-level Memorization}
\label{sec:label_shuffling}

We now examine how the distribution
of information measures across networks
change when they memorize individual examples.
Following our original hypotheses (Table \ref{tb:relationships}; \ref{item:hyp_2}),
we expect such networks to display high
\divi\ and \divii\ diversity, and thus 
high entropy and low MI.

We perform the analysis for \memi\ memorization
on the standard datasets of 
\memiSynthSetI~\citep{lecun1998mnist} and 
\memiSynthSetII~\citep{maas2011imdb} on 
a 3-layered MLP and 
DistilBERT-base, respectively.
In order to study how the 
diversity of neurons 
changes with
increasing \memi\ memorization,
we induce varying levels of label noise
by randomly shuffling a fraction of 
training examples' target labels 
(denoted by a parameter $\beta$).
We then analyze these trained networks
on the original validation set.

\begin{figure*}[t]
    \centering
    \begin{subfigure}[t]{0.49\linewidth}
        \centering
        \includegraphics[width=\textwidth]{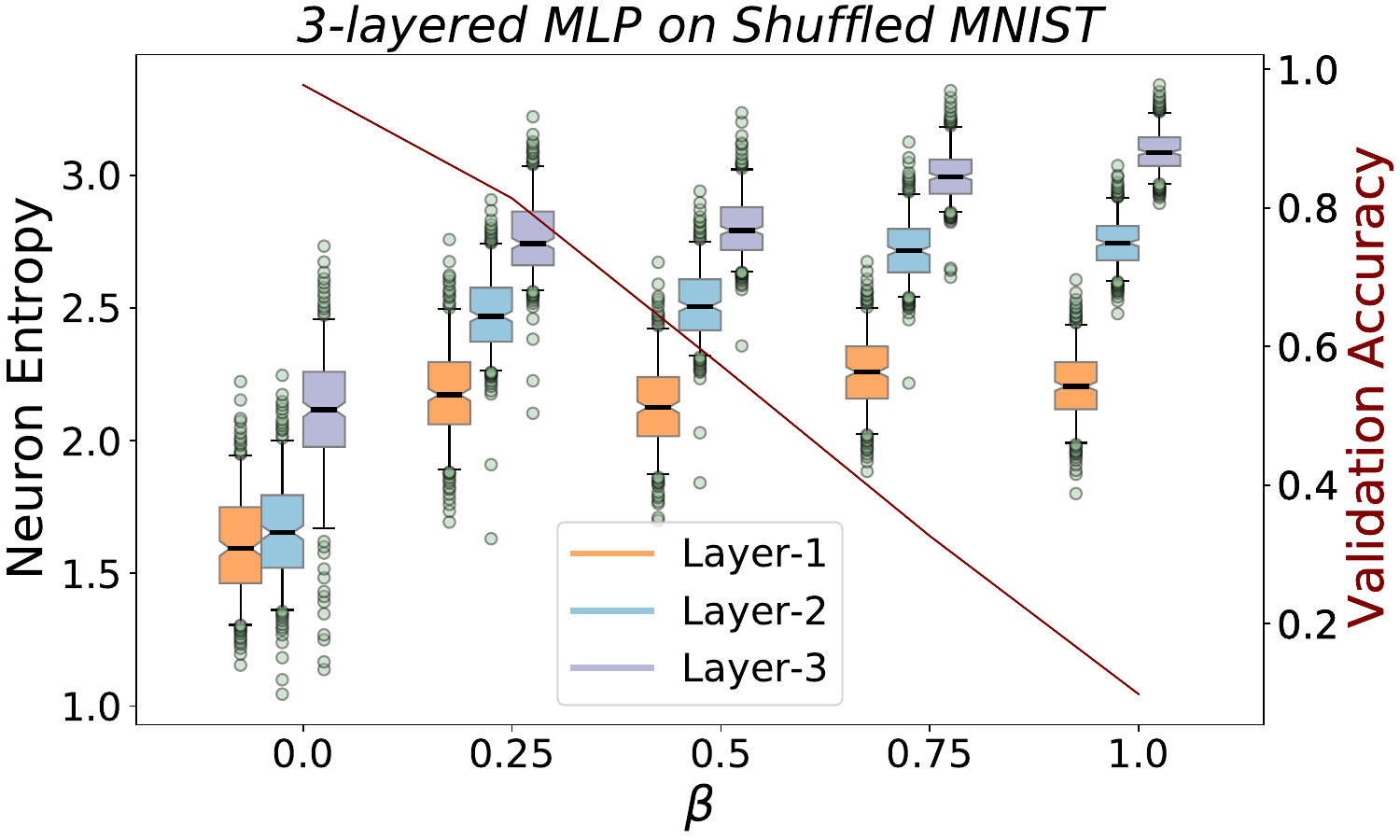}
    \end{subfigure}
    \hfill
    \begin{subfigure}[t]{0.49\linewidth}
        \centering
        \includegraphics[width=\textwidth]{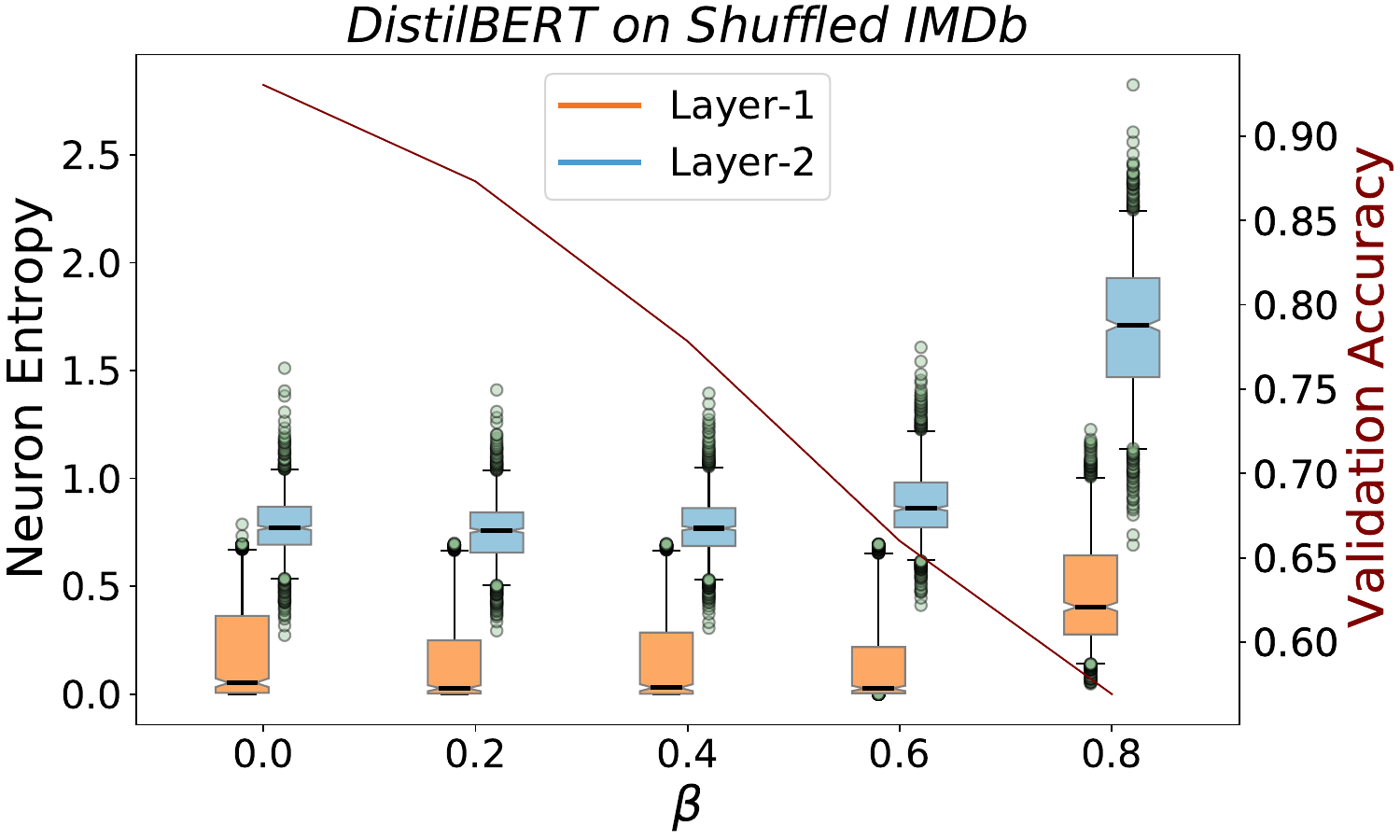}
    \end{subfigure}
    \caption{
    Entropy across neuron activations 
    increases with greater \memi\ memorization
    ($\uparrow \beta$).
    \label{fig:label_shuffling_entropy}
    }
\end{figure*}

\paragraph{Results:}
First, 
we note that model performance 
on the 
validation set 
decreases with increased label shuffling,
validating an increase in \memi\ memorization (Figure~\ref{fig:label_shuffling_entropy}).
Interestingly, this dip in validation accuracy 
is accompanied with a consistent rise in 
entropy across the neurons. 
For MLP networks trained on \memiSynthSetI\ (left),
we see a distinct rise in entropy
even with a small amount of label shuffling ($\beta = 0.25$), 
followed by a steady increase (layers 2 and 3)
or no change (layer 1) in entropy.
A dissimilar trend is seen for DistilBERT
fine-tuned on \memiSynthSetII\ (right):
a distinct rise 
for high values of $\beta$ 
and a consistent value for low or no
label shuffling.
While our hypothesis holds true in both settings,
we speculate the difference between them
is due to the pre-trained initialization of DistilBERT,
which has been shown to act as an 
implicit regularization 
during fine-tuning
\citep{tu2020bert}.
That is, here, DistilBERT might be learning 
task-relevant information 
despite some amount of label noise
(note that this is not evident through validation performance alone).

\begin{figure*}[t]
    \centering
    \begin{subfigure}[t]{0.48\linewidth}
        \centering
        \includegraphics[width=\textwidth]{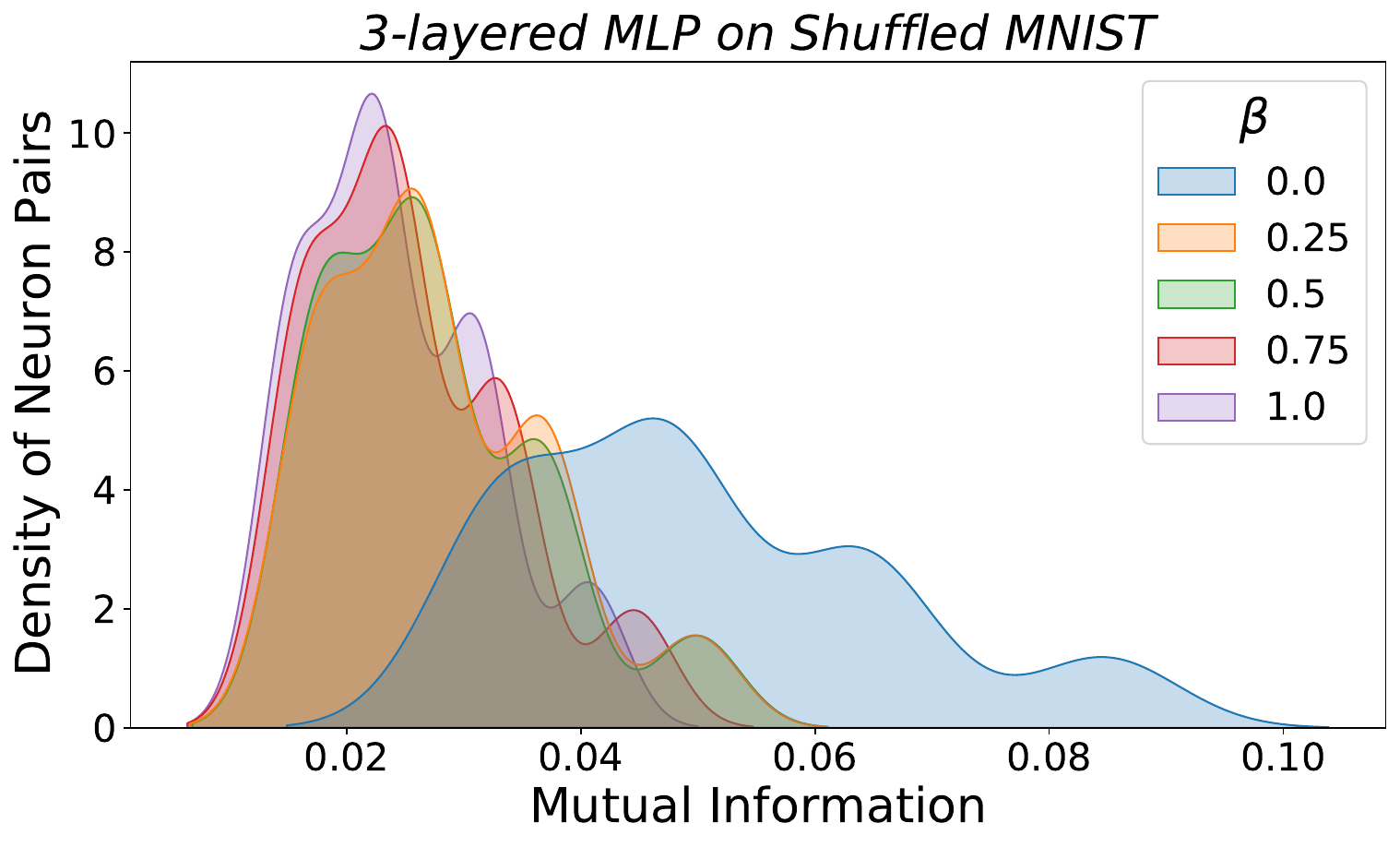}
    \end{subfigure}
    \hfill
    \begin{subfigure}[t]{0.48\linewidth}
        \centering
        \includegraphics[width=\textwidth]{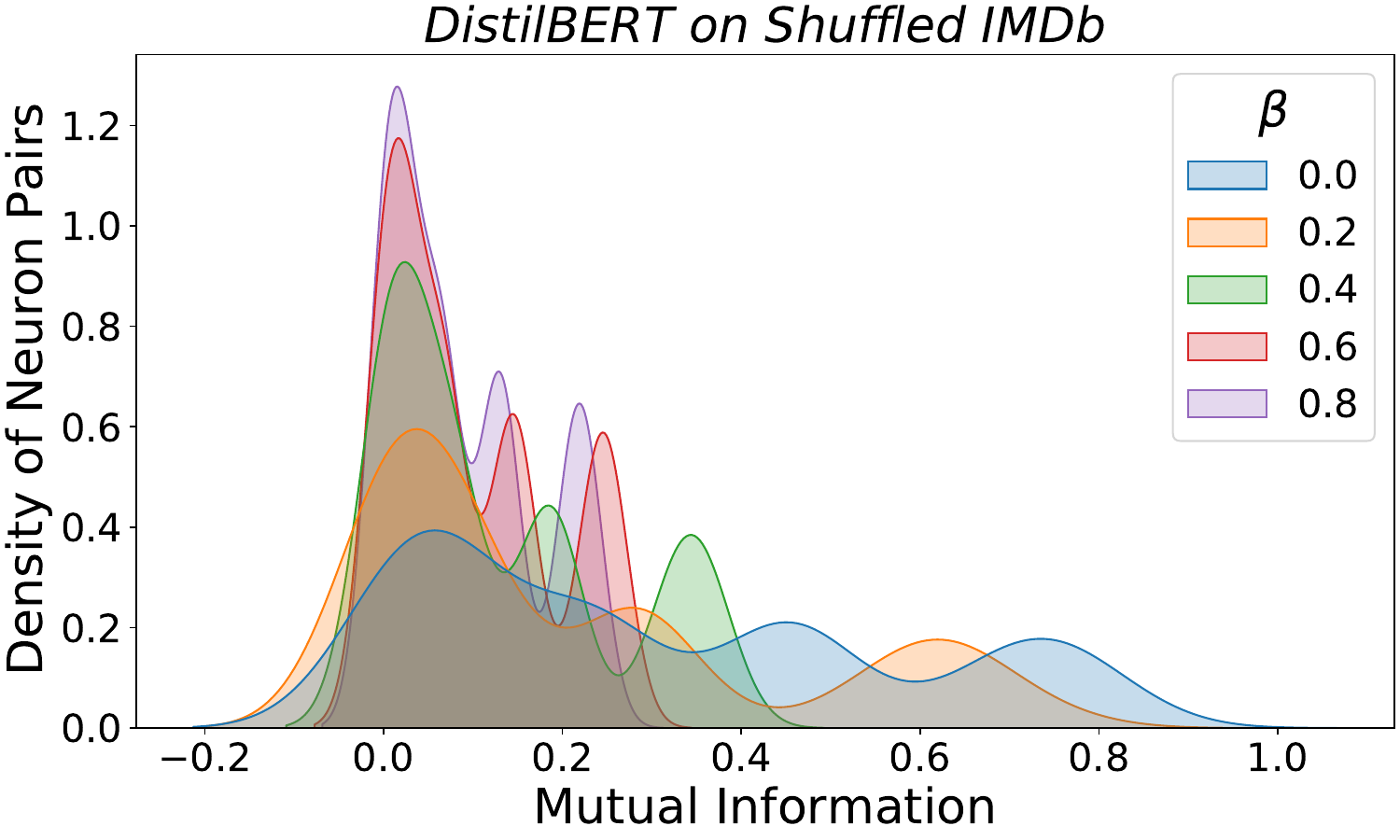}
        \label{sfig:shuffled_imdb_mi}
    \end{subfigure}
    \caption{
    Networks that show higher \memi\ memorization
    ($\uparrow \beta$)
    have high density of neuron pairs
    for lower MI values.
    Here, MI is computed across the first
    layer for both the networks.
    \label{fig:label_shuffling_mi}
    }
\end{figure*}

Our hypothesis for the relation between
\memi\ memorization and MI
is 
supported by
Figure~\ref{fig:label_shuffling_mi}.
In both settings,
networks trained on higher $\beta$ values
consist of neuron pairs
that show low values of MI
(left side of the plots).
In line with the previous observations, 
we find that MLPs trained on some amount
of label noise (any $\beta$ > 0.00) on MNIST (left sub-plot)
have a higher 
density of neuron pairs concentrated
at low values of MI.
Meanwhile, for DistilBERT on \memiSynthSetII\ (right sub-plot), 
we observe  
that neuron pair density gradually shifts towards
lower values of MI with increasing $\beta$.%
\footnote{
Although MI values remain non-negative throughout,
the x-axis in our density plots might show negative values
as an artifact of fitting a Gaussian mixture model.
}
\section{Model Selection}
\label{sec:model_selection}

\begin{table}[ht]
\centering
\caption{
We measure the correlation (Kendall's $\tau$) 
between 
model rankings
based on their generalization 
as estimated 
through extrinsic metrics on labeled test sets
and those obtained via information measures.
Note that 
$\tau$ can range from -1.0 (perfect disagreement) 
to 1.0 (perfect agreement).
}
\label{tb:model_selection}

\begin{tabular}{l|crrrrrr}
\toprule
             & \multicolumn{1}{c|}{\cellcolor[HTML]{FAEDE0}\begin{tabular}[c]{@{}c@{}}Sentiment\\ Adjectives\end{tabular}} & \multicolumn{1}{c|}{\cellcolor[HTML]{FAEDE0}\begin{tabular}[c]{@{}c@{}}Colored\\ MNIST\end{tabular}} & \multicolumn{3}{c|}{\cellcolor[HTML]{FAEDE0}Bias-in-Bios}                                                                                                                                                                      & \multicolumn{1}{c|}{\cellcolor[HTML]{ECF3E9}\begin{tabular}[c]{@{}c@{}}Shuffled\\ IMDb\end{tabular}} & \multicolumn{1}{c}{\cellcolor[HTML]{ECF3E9}\begin{tabular}[c]{@{}c@{}}Shuffled\\ MNIST\end{tabular}} \\ \cmidrule{2-8}
             & \multicolumn{1}{l|}{\begin{tabular}[c]{@{}l@{}}Validation\\ Accuracy\end{tabular}}                                & \multicolumn{1}{l|}{\begin{tabular}[c]{@{}l@{}}Validation\\ Accuracy\end{tabular}}                         & \multicolumn{1}{l|}{\begin{tabular}[c]{@{}l@{}}Comp-\\ ression\end{tabular}} & \multicolumn{1}{l|}{\begin{tabular}[c]{@{}l@{}}TPR\\ Gap\end{tabular}} & \multicolumn{1}{l|}{\begin{tabular}[c]{@{}l@{}}Suff.\\ Gap\end{tabular}} & \multicolumn{1}{l|}{\begin{tabular}[c]{@{}l@{}}Validation\\ Accuracy\end{tabular}}                         & \multicolumn{1}{l}{\begin{tabular}[c]{@{}l@{}}Validation\\ Accuracy\end{tabular}}                          \\ \midrule
Mean Entropy & \multicolumn{1}{c|}{\cellcolor[HTML]{C6C6C6}0.80}                                                           & \multicolumn{1}{c|}{\cellcolor[HTML]{B7B7B7}1.00}                                                    & \multicolumn{1}{c|}{\cellcolor[HTML]{DEDEDE}0.47}                                                & \multicolumn{1}{c|}{\cellcolor[HTML]{F1F1F1}0.20}                                          & \multicolumn{1}{c|}{\cellcolor[HTML]{F1F1F1}0.20}                        & \multicolumn{1}{c|}{\cellcolor[HTML]{D4D4D4}0.60}                                                    & \multicolumn{1}{c}{\cellcolor[HTML]{B7B7B7}1.00}                                                                         \\ \hline
Mean MI      & \multicolumn{1}{c|}{\cellcolor[HTML]{C6C6C6}0.80}                                                           & \multicolumn{1}{c|}{\cellcolor[HTML]{B7B7B7}1.00}                                                    & \multicolumn{1}{c|}{\cellcolor[HTML]{D4D4D4}0.60}                                                & \multicolumn{1}{c|}{\cellcolor[HTML]{FBFBFB}0.07}                                          & \multicolumn{1}{c|}{\cellcolor[HTML]{E7E7E7}0.33}                        & \multicolumn{1}{c|}{\cellcolor[HTML]{C6C6C6}0.80}                                                    & \multicolumn{1}{c}{\cellcolor[HTML]{B7B7B7}1.00}                                                                         \\ \hline
\end{tabular}
\end{table}


In the previous sections, we have seen that 
studying information organization
through the presented measures
allow us to qualitatively distinguish networks
with different generalizing behaviors. 
A natural application of our 
findings is the problem of model selection: 
given a list of models, 
rank them based on their generalizability.
To demonstrate
the utility of our insights,
we compare the correlations 
between rankings obtained 
through our information-theoretic measures 
(which do no require labeled data)
and the 
generalization ability of the model on a labeled held-out set.

We consider the same tasks and networks
as discussed in the prior sections,
and compute the rankings
using (i) extrinsic evaluation metrics
defined for the task
(such as validation accuracy for 
\memiiSynthSetI\ and 
compression for \memiiNatSetI), 
(ii) the mean of entropy values, and 
(iii) the mean of MI values 
computed for the same networks.
We then compute the Kendall rank correlation 
coefficient, $\tau$, between these rankings
(between (i) \& (ii), and (i) \& (iii))
to evaluate the agreement amongst them.

We observe high 
correlation values for all the comparisons~(Table~\ref{tb:model_selection}).
Particularly high correlations are observed for setups
with synthetically induced spurious correlations
(\cref{subsec:semi_synthetic_memii}) and
shuffled labels (\cref{sec:label_shuffling}),
with rankings on \memiiSynthSetI\ being perfectly correlated.
Correlations on \memiiNatSetI\ are positive but lower, likely
due to the more nuanced setup, 
where the memorization is less pronounced
and extrinsic metrics are weakly correlated
even among themselves 
\citep[\cref{subsec:biasinbias_model_rankings}; ][]{orgad2022debiasing}.
These positive correlations are important
because---unlike the other metrics
across which the correlations are computed---the information measures are purely
intrinsic to the model and do not assume
access to any OOD data.
We perform an additional comparative discussion
with standard conventional methods for model selection
in \cref{subsec:model_selection_comparison}.
\section{Related Work}

A large body of work 
aims at measuring and quantifying generalization,
especially in out-of-distribution (OOD) scenarios~\citep{david2010ood, hendrycks2021ood, wang2021ood}.
The most common approach 
is to curate and label a set of examples 
to evaluate if 
networks exploit 
certain heuristics or shortcuts~\citep{lapuschkin2019challenge, zhao2018generating}.
Several past studies create such sets
spanning different  domains and tasks to shed light
on common failure modes in both 
the trained models, 
and the datasets used to train them. 
A body of such work exists for several tasks across 
vision~\citep{russakovsky2015imagenet, hendrycks2019benchmarking,  hendrycks2021ood, hendrycks2021nae}
and NLP~\citep{mccoy-etal-2019-right, naik2021stress, ravichander-etal-2021-noiseqa, kim2020cogs}.

Closely related to the motivations for our work,
past work has attempted to evaluate models 
using techniques that go beyond extrinsic
evaluation.
Training dynamics have been explored to assess
the role of individual examples in training sets~\citep{swayamdipta2020cartography}
and how specific knowledge features are temporally picked during training 
\citep{saphra-lopez-2019-understanding}.
Recent work has noted that frequent spurious artifacts are learnt
prior to general patterns during training~\citep{tanzer2021bertmem},
in turn followed by memorization of individual examples
\citep{arpit2018memorization}.
Closely sharing our motivations of using information-theoretic viewpoints
for intrinsic evaluation over network activations,
past work has investigated \emph{probing} or \emph{diagnostic} classifiers
\citep{ettinger-etal-2016-probing,adi2017probing, belinkov2017probing, Hupkes2018VisualisationA}.
Researchers have further extended this paradigm
to analyze 
the role of individual neurons
\citep{dalvi2019neurons, durrani2020neurons, Bau2017NetworkDQ,bau2020units}, although this approach may fail to identify causal roles~\citep{antverg2021neurons,belinkov2022probing}.
Other work using information theory 
to study neural networks
has focused on their learning process 
through the lens of the information bottleneck principle
\citep{tishby99information},
categorizing learning into distinct phases 
\citep{ziv2017ib, saxe2018ib}
and obtaining generalization bounds~\citep{tishby2015ib}.
Follow-up work has made use of such measures
to regularize training for robustness 
\citep{wang2021infobert} and low-resource learning
\citep{mahabadi2021ib}.
\section{Limitations and Future Directions}
\label{sec:limitations}

Below, we describe some of the limitations of our work and discuss future research directions.

\paragraph{Comparative Nature of Observations}
\label{para:comparison}
The current findings and insights 
derived from the information measures
are comparative in nature that
could be a limitation
when being applied for practical
use cases.
In order to assess some given models
using the information measures described herein,
we must a-priori know at least one of two things:
(i) one of the given models that generalizes
well, so that the rest could be bench-marked against it,
or (ii) the kind of memorization 
that the models trained on the dataset 
are expected to possess, so that we could make
a comparison between the given models. 
Further, one may want to compare
models that do not belong to the same model family,
architecture and hyperparameter set.
In such cases, 
values from our
information measures might not be directly comparable
across these different models.
We design a simple experiment to study this
in \cref{subsec:capacity_effect} where
we compute our measures for models 
with varying capacity. We note that values
for networks with different capacities 
lie on different scales and hence
are not directly comparable.

\paragraph{Scaling to Larger Models}
\label{para:scale}
While the analysis presented in this work
is performed for small to moderately large
networks like MLPs, RoBERTa, and ResNet---for
whom our hypothesized trend holds
consistently---more research is needed to 
study the scaling behavior of these
information measures as a function of data 
and model size~\citep{rosenfield2020scaling,Kaplan2020ScalingLF}.

\paragraph{Practical Applications.}
In this work, we show the utility of our observations for the preliminary 
use case of model selection. More research is required to investigate the usefulness of our observations in other scenarios. 
%
One viable direction to explore is the problem of 
\emph{OOD detection}~\citep{arora2021ood}---deciding whether a specific data point is OOD---by computing point-wise versions of the information measures.  
Another case where the proposed information measures
could also be useful is \emph{regularizing models},
where regularizing the MI/entropy values to a certain a band of values
might yield more generalizing models.
Such regularization can also be coupled with
our understanding of training dynamics from prior work
\citep{tanzer2021bertmem} that has identified
training stages where particular forms of 
memorization is seen to exist.
Our understanding of \emph{training dynamics}, in itself,
could be enhanced by studying the progression of neuron 
diversity across training steps and hence noting the
generalization patterns that emerge.
\section{Conclusion}
\label{sec:conclusion}

In this work, we have taken a step towards
identifying generalization behavior of 
neural network models based on their intrinsic
activation patterns.
We presented information-theoretic measures
that allow us to distinguish between models 
that show two kinds of memorization:
those that pick up surface-level spurious correlations
(\memii\ memorization) and 
those that overfit on individual training instances
(\memi\ memorization).
Through investigations spanning multiple natural language and vision tasks,
we corroborated our hypothesis that such memorization
is reflected in diversity across neural activations,
and hence the defined information measures that quantify them.
Finally, we demonstrated a potential application of this framework
for model selection.

\section*{Acknowledgments}
We are grateful to 
the Technion CS NLP group and others at the Technion---particularly, 
Mor Ventura, Michael Toker, Hadas Orgad, Reda Igbaria, Zach Bamberger,
Adir Rahamim, Anja Reusch, and Gail Weiss---for the insightful discussions that shaped this work.
RB would like to extend his gratitude to his
dorm-mates and friends---%
Atulya, Josh, Cornelius, David, Krist\'{o}f,
Pratibha, Ajay, Navdeep---%
for being a constant source of home
during his time at the Technion.
RB would also like to thank the support staff at Delhi Technological University and the Technion for their administrative support.
We also thank the anonymous reviewers and area chairs during the
review process at NeurIPS 2022 for their careful analysis of our work.
This research was supported by the Israel Science Foundation (grant No.\ 448/20), an Azrieli Foundation Early Career Faculty Fellowship, and an AI Alignment grant from Open Philanthropy.

\bibliography{neurips_2022}

\begin{thebibliography}{58}
\providecommand{\natexlab}[1]{#1}
\providecommand{\url}[1]{\texttt{#1}}
\expandafter\ifx\csname urlstyle\endcsname\relax
  \providecommand{\doi}[1]{doi: #1}\else
  \providecommand{\doi}{doi: \begingroup \urlstyle{rm}\Url}\fi

\bibitem[Adi et~al.(2017)Adi, Kermany, Belinkov, Lavi, and
  Goldberg]{adi2017probing}
Adi, Y., Kermany, E., Belinkov, Y., Lavi, O., and Goldberg, Y.
\newblock Fine-grained analysis of sentence embeddings using auxiliary
  prediction tasks.
\newblock In \emph{5th International Conference on Learning Representations,
  {ICLR} 2017, Toulon, France, April 24-26, 2017, Conference Track
  Proceedings}. OpenReview.net, 2017.
\newblock URL \url{https://openreview.net/forum?id=BJh6Ztuxl}.

\bibitem[Antverg \& Belinkov(2022)Antverg and Belinkov]{antverg2021neurons}
Antverg, O. and Belinkov, Y.
\newblock On the pitfalls of analyzing individual neurons in language models.
\newblock In \emph{International Conference on Learning Representations}, 2022.
\newblock URL \url{https://openreview.net/forum?id=8uz0EWPQIMu}.

\bibitem[Arjovsky et~al.(2019)Arjovsky, Bottou, Gulrajani, and
  Lopez{-}Paz]{arjovsky2019irm}
Arjovsky, M., Bottou, L., Gulrajani, I., and Lopez{-}Paz, D.
\newblock Invariant risk minimization.
\newblock \emph{CoRR}, abs/1907.02893, 2019.
\newblock URL \url{http://arxiv.org/abs/1907.02893}.

\bibitem[Arora et~al.(2021)Arora, Huang, and He]{arora2021ood}
Arora, U., Huang, W., and He, H.
\newblock Types of out-of-distribution texts and how to detect them.
\newblock In Moens, M., Huang, X., Specia, L., and Yih, S.~W. (eds.),
  \emph{Proceedings of the 2021 Conference on Empirical Methods in Natural
  Language Processing, {EMNLP} 2021, Virtual Event / Punta Cana, Dominican
  Republic, 7-11 November, 2021}, pp.\  10687--10701. Association for
  Computational Linguistics, 2021.
\newblock \doi{10.18653/v1/2021.emnlp-main.835}.
\newblock URL \url{https://doi.org/10.18653/v1/2021.emnlp-main.835}.

\bibitem[Arpit et~al.(2017)Arpit, Jastrzebski, Ballas, Krueger, Bengio, Kanwal,
  Maharaj, Fischer, Courville, Bengio, and
  Lacoste{-}Julien]{arpit2018memorization}
Arpit, D., Jastrzebski, S., Ballas, N., Krueger, D., Bengio, E., Kanwal, M.~S.,
  Maharaj, T., Fischer, A., Courville, A.~C., Bengio, Y., and Lacoste{-}Julien,
  S.
\newblock A closer look at memorization in deep networks.
\newblock In Precup, D. and Teh, Y.~W. (eds.), \emph{Proceedings of the 34th
  International Conference on Machine Learning, {ICML} 2017, Sydney, NSW,
  Australia, 6-11 August 2017}, volume~70 of \emph{Proceedings of Machine
  Learning Research}, pp.\  233--242. {PMLR}, 2017.
\newblock URL \url{http://proceedings.mlr.press/v70/arpit17a.html}.

\bibitem[Bau et~al.(2017)Bau, Zhou, Khosla, Oliva, and
  Torralba]{Bau2017NetworkDQ}
Bau, D., Zhou, B., Khosla, A., Oliva, A., and Torralba, A.
\newblock Network dissection: Quantifying interpretability of deep visual
  representations.
\newblock \emph{2017 IEEE Conference on Computer Vision and Pattern Recognition
  (CVPR)}, pp.\  3319--3327, 2017.

\bibitem[Bau et~al.(2020)Bau, Zhu, Strobelt, Lapedriza, Zhou, and
  Torralba]{bau2020units}
Bau, D., Zhu, J.-Y., Strobelt, H., Lapedriza, A., Zhou, B., and Torralba, A.
\newblock Understanding the role of individual units in a deep neural network.
\newblock \emph{Proceedings of the National Academy of Sciences}, 2020.
\newblock ISSN 0027-8424.
\newblock \doi{10.1073/pnas.1907375117}.
\newblock URL \url{https://www.pnas.org/content/early/2020/08/31/1907375117}.

\bibitem[Beaudry \& Renner(2012)Beaudry and Renner]{beaudry2012dpi}
Beaudry, N.~J. and Renner, R.
\newblock An intuitive proof of the data processing inequality.
\newblock \emph{Quantum Inf. Comput.}, 12\penalty0 (5-6):\penalty0 432--441,
  2012.
\newblock \doi{10.26421/QIC12.5-6-4}.
\newblock URL \url{https://doi.org/10.26421/QIC12.5-6-4}.

\bibitem[Belghazi et~al.(2018)Belghazi, Baratin, Rajeswar, Ozair, Bengio,
  Hjelm, and Courville]{belghazi2018mi}
Belghazi, M.~I., Baratin, A., Rajeswar, S., Ozair, S., Bengio, Y., Hjelm,
  R.~D., and Courville, A.~C.
\newblock Mutual information neural estimation.
\newblock In Dy, J.~G. and Krause, A. (eds.), \emph{Proceedings of the 35th
  International Conference on Machine Learning, {ICML} 2018,
  Stockholmsm{\"{a}}ssan, Stockholm, Sweden, July 10-15, 2018}, volume~80 of
  \emph{Proceedings of Machine Learning Research}, pp.\  530--539. {PMLR},
  2018.
\newblock URL \url{http://proceedings.mlr.press/v80/belghazi18a.html}.

\bibitem[Belinkov(2022)]{belinkov2022probing}
Belinkov, Y.
\newblock Probing classifiers: Promises, shortcomings, and advances.
\newblock \emph{Comput. Linguistics}, 48\penalty0 (1):\penalty0 207--219, 2022.
\newblock \doi{10.1162/coli\_a\_00422}.
\newblock URL \url{https://doi.org/10.1162/coli\_a\_00422}.

\bibitem[Belinkov et~al.(2017)Belinkov, Durrani, Dalvi, Sajjad, and
  Glass]{belinkov2017probing}
Belinkov, Y., Durrani, N., Dalvi, F., Sajjad, H., and Glass, J.~R.
\newblock What do neural machine translation models learn about morphology?
\newblock In Barzilay, R. and Kan, M. (eds.), \emph{Proceedings of the 55th
  Annual Meeting of the Association for Computational Linguistics, {ACL} 2017,
  Vancouver, Canada, July 30 - August 4, Volume 1: Long Papers}, pp.\
  861--872. Association for Computational Linguistics, 2017.
\newblock \doi{10.18653/v1/P17-1080}.
\newblock URL \url{https://doi.org/10.18653/v1/P17-1080}.

\bibitem[Ben{-}David et~al.(2010)Ben{-}David, Blitzer, Crammer, Kulesza,
  Pereira, and Vaughan]{david2010ood}
Ben{-}David, S., Blitzer, J., Crammer, K., Kulesza, A., Pereira, F., and
  Vaughan, J.~W.
\newblock A theory of learning from different domains.
\newblock \emph{Mach. Learn.}, 79\penalty0 (1-2):\penalty0 151--175, 2010.
\newblock \doi{10.1007/s10994-009-5152-4}.
\newblock URL \url{https://doi.org/10.1007/s10994-009-5152-4}.

\bibitem[Cheng et~al.(2020)Cheng, Hao, Dai, Liu, Gan, and Carin]{cheng2020club}
Cheng, P., Hao, W., Dai, S., Liu, J., Gan, Z., and Carin, L.
\newblock {CLUB:} {A} contrastive log-ratio upper bound of mutual information.
\newblock In \emph{Proceedings of the 37th International Conference on Machine
  Learning, {ICML} 2020, 13-18 July 2020, Virtual Event}, volume 119 of
  \emph{Proceedings of Machine Learning Research}, pp.\  1779--1788. {PMLR},
  2020.
\newblock URL \url{http://proceedings.mlr.press/v119/cheng20b.html}.

\bibitem[Dalvi et~al.(2019)Dalvi, Durrani, Sajjad, Belinkov, Bau, and
  Glass]{dalvi2019neurons}
Dalvi, F., Durrani, N., Sajjad, H., Belinkov, Y., Bau, A., and Glass, J.~R.
\newblock What is one grain of sand in the desert? analyzing individual neurons
  in deep {NLP} models.
\newblock In \emph{The Thirty-Third {AAAI} Conference on Artificial
  Intelligence, {AAAI} 2019, The Thirty-First Innovative Applications of
  Artificial Intelligence Conference, {IAAI} 2019, The Ninth {AAAI} Symposium
  on Educational Advances in Artificial Intelligence, {EAAI} 2019, Honolulu,
  Hawaii, USA, January 27 - February 1, 2019}, pp.\  6309--6317. {AAAI} Press,
  2019.
\newblock \doi{10.1609/aaai.v33i01.33016309}.
\newblock URL \url{https://doi.org/10.1609/aaai.v33i01.33016309}.

\bibitem[Darbellay \& Vajda(1999)Darbellay and Vajda]{darbellay1999info}
Darbellay, G. and Vajda, I.
\newblock Estimation of the information by an adaptive partitioning of the
  observation space.
\newblock \emph{IEEE Transactions on Information Theory}, 45\penalty0
  (4):\penalty0 1315--1321, 1999.
\newblock \doi{10.1109/18.761290}.

\bibitem[De{-}Arteaga et~al.(2019)De{-}Arteaga, Romanov, Wallach, Chayes,
  Borgs, Chouldechova, Geyik, Kenthapadi, and Kalai]{de-arteaga2019biasinbios}
De{-}Arteaga, M., Romanov, A., Wallach, H.~M., Chayes, J.~T., Borgs, C.,
  Chouldechova, A., Geyik, S.~C., Kenthapadi, K., and Kalai, A.~T.
\newblock Bias in bios: {A} case study of semantic representation bias in a
  high-stakes setting.
\newblock In danah boyd and Morgenstern, J.~H. (eds.), \emph{Proceedings of the
  Conference on Fairness, Accountability, and Transparency, FAT* 2019, Atlanta,
  GA, USA, January 29-31, 2019}, pp.\  120--128. {ACM}, 2019.
\newblock \doi{10.1145/3287560.3287572}.
\newblock URL \url{https://doi.org/10.1145/3287560.3287572}.

\bibitem[Durrani et~al.(2020)Durrani, Sajjad, Dalvi, and
  Belinkov]{durrani2020neurons}
Durrani, N., Sajjad, H., Dalvi, F., and Belinkov, Y.
\newblock Analyzing individual neurons in pre-trained language models.
\newblock In Webber, B., Cohn, T., He, Y., and Liu, Y. (eds.),
  \emph{Proceedings of the 2020 Conference on Empirical Methods in Natural
  Language Processing, {EMNLP} 2020, Online, November 16-20, 2020}, pp.\
  4865--4880. Association for Computational Linguistics, 2020.
\newblock \doi{10.18653/v1/2020.emnlp-main.395}.
\newblock URL \url{https://doi.org/10.18653/v1/2020.emnlp-main.395}.

\bibitem[Ettinger et~al.(2016)Ettinger, Elgohary, and
  Resnik]{ettinger-etal-2016-probing}
Ettinger, A., Elgohary, A., and Resnik, P.
\newblock Probing for semantic evidence of composition by means of simple
  classification tasks.
\newblock In \emph{Proceedings of the 1st Workshop on Evaluating Vector-Space
  Representations for {NLP}}, pp.\  134--139, Berlin, Germany, August 2016.
  Association for Computational Linguistics.
\newblock \doi{10.18653/v1/W16-2524}.
\newblock URL \url{https://aclanthology.org/W16-2524}.

\bibitem[Geirhos et~al.(2019)Geirhos, Rubisch, Michaelis, Bethge, Wichmann, and
  Brendel]{geirhos2019cnns}
Geirhos, R., Rubisch, P., Michaelis, C., Bethge, M., Wichmann, F.~A., and
  Brendel, W.
\newblock Imagenet-trained cnns are biased towards texture; increasing shape
  bias improves accuracy and robustness.
\newblock In \emph{7th International Conference on Learning Representations,
  {ICLR} 2019, New Orleans, LA, USA, May 6-9, 2019}. OpenReview.net, 2019.
\newblock URL \url{https://openreview.net/forum?id=Bygh9j09KX}.

\bibitem[Geirhos et~al.(2020)Geirhos, Jacobsen, Michaelis, Zemel, Brendel,
  Bethge, and Wichmann]{geirhos2020shortcut}
Geirhos, R., Jacobsen, J.-H., Michaelis, C., Zemel, R., Brendel, W., Bethge,
  M., and Wichmann, F.~A.
\newblock Shortcut learning in deep neural networks.
\newblock \emph{Nature Machine Intelligence}, 2\penalty0 (11):\penalty0
  665--673, 2020.

\bibitem[He et~al.(2015)He, Zhang, Ren, and Sun]{he2015resnet}
He, K., Zhang, X., Ren, S., and Sun, J.
\newblock Deep residual learning for image recognition.
\newblock \emph{CoRR}, abs/1512.03385, 2015.
\newblock URL \url{http://arxiv.org/abs/1512.03385}.

\bibitem[Hendrycks \& Dietterich(2019)Hendrycks and
  Dietterich]{hendrycks2019benchmarking}
Hendrycks, D. and Dietterich, T.~G.
\newblock Benchmarking neural network robustness to common corruptions and
  perturbations.
\newblock In \emph{7th International Conference on Learning Representations,
  {ICLR} 2019, New Orleans, LA, USA, May 6-9, 2019}. OpenReview.net, 2019.
\newblock URL \url{https://openreview.net/forum?id=HJz6tiCqYm}.

\bibitem[Hendrycks et~al.(2021{\natexlab{a}})Hendrycks, Basart, Mu, Kadavath,
  Wang, Dorundo, Desai, Zhu, Parajuli, Guo, Song, Steinhardt, and
  Gilmer]{hendrycks2021ood}
Hendrycks, D., Basart, S., Mu, N., Kadavath, S., Wang, F., Dorundo, E., Desai,
  R., Zhu, T., Parajuli, S., Guo, M., Song, D., Steinhardt, J., and Gilmer, J.
\newblock The many faces of robustness: {A} critical analysis of
  out-of-distribution generalization.
\newblock In \emph{2021 {IEEE/CVF} International Conference on Computer Vision,
  {ICCV} 2021, Montreal, QC, Canada, October 10-17, 2021}, pp.\  8320--8329.
  {IEEE}, 2021{\natexlab{a}}.
\newblock \doi{10.1109/ICCV48922.2021.00823}.
\newblock URL \url{https://doi.org/10.1109/ICCV48922.2021.00823}.

\bibitem[Hendrycks et~al.(2021{\natexlab{b}})Hendrycks, Zhao, Basart,
  Steinhardt, and Song]{hendrycks2021nae}
Hendrycks, D., Zhao, K., Basart, S., Steinhardt, J., and Song, D.
\newblock Natural adversarial examples.
\newblock In \emph{{IEEE} Conference on Computer Vision and Pattern
  Recognition, {CVPR} 2021, virtual, June 19-25, 2021}, pp.\  15262--15271.
  Computer Vision Foundation / {IEEE}, 2021{\natexlab{b}}.
\newblock URL
  \url{https://openaccess.thecvf.com/content/CVPR2021/html/Hendrycks\_Natural\_Adversarial\_Examples\_CVPR\_2021\_paper.html}.

\bibitem[Hupkes et~al.(2018)Hupkes, Veldhoen, and
  Zuidema]{Hupkes2018VisualisationA}
Hupkes, D., Veldhoen, S., and Zuidema, W.~H.
\newblock Visualisation and 'diagnostic classifiers' reveal how recurrent and
  recursive neural networks process hierarchical structure.
\newblock \emph{J. Artif. Intell. Res.}, 61:\penalty0 907--926, 2018.

\bibitem[Jiang et~al.(2019)Jiang, Neyshabur, Mobahi, Krishnan, and
  Bengio]{yiding2019fantastic}
Jiang, Y., Neyshabur, B., Mobahi, H., Krishnan, D., and Bengio, S.
\newblock Fantastic generalization measures and where to find them, 2019.
\newblock URL \url{https://arxiv.org/abs/1912.02178}.

\bibitem[Kaplan et~al.(2020)Kaplan, McCandlish, Henighan, Brown, Chess, Child,
  Gray, Radford, Wu, and Amodei]{Kaplan2020ScalingLF}
Kaplan, J., McCandlish, S., Henighan, T.~J., Brown, T.~B., Chess, B., Child,
  R., Gray, S., Radford, A., Wu, J., and Amodei, D.
\newblock Scaling laws for neural language models.
\newblock \emph{ArXiv}, abs/2001.08361, 2020.

\bibitem[Kim \& Linzen(2020)Kim and Linzen]{kim2020cogs}
Kim, N. and Linzen, T.
\newblock {COGS:} {A} compositional generalization challenge based on semantic
  interpretation.
\newblock In Webber, B., Cohn, T., He, Y., and Liu, Y. (eds.),
  \emph{Proceedings of the 2020 Conference on Empirical Methods in Natural
  Language Processing, {EMNLP} 2020, Online, November 16-20, 2020}, pp.\
  9087--9105. Association for Computational Linguistics, 2020.
\newblock \doi{10.18653/v1/2020.emnlp-main.731}.
\newblock URL \url{https://doi.org/10.18653/v1/2020.emnlp-main.731}.

\bibitem[Kraskov et~al.(2004)Kraskov, St\"ogbauer, and
  Grassberger]{kraskov2004mi}
Kraskov, A., St\"ogbauer, H., and Grassberger, P.
\newblock Estimating mutual information.
\newblock \emph{Phys. Rev. E}, 69:\penalty0 066138, Jun 2004.
\newblock \doi{10.1103/PhysRevE.69.066138}.
\newblock URL \url{https://link.aps.org/doi/10.1103/PhysRevE.69.066138}.

\bibitem[Lapuschkin et~al.(2019)Lapuschkin, W{\"{a}}ldchen, Binder, Montavon,
  Samek, and M{\"{u}}ller]{lapuschkin2019challenge}
Lapuschkin, S., W{\"{a}}ldchen, S., Binder, A., Montavon, G., Samek, W., and
  M{\"{u}}ller, K.
\newblock Unmasking clever hans predictors and assessing what machines really
  learn.
\newblock \emph{CoRR}, abs/1902.10178, 2019.
\newblock URL \url{http://arxiv.org/abs/1902.10178}.

\bibitem[Lecun et~al.(1998)Lecun, Bottou, Bengio, and Haffner]{lecun1998mnist}
Lecun, Y., Bottou, L., Bengio, Y., and Haffner, P.
\newblock Gradient-based learning applied to document recognition.
\newblock \emph{Proceedings of the IEEE}, 86\penalty0 (11):\penalty0
  2278--2324, 1998.
\newblock \doi{10.1109/5.726791}.

\bibitem[Lee et~al.(2020)Lee, Lee, Hwang, Yang, and Choi]{lee2020complexity}
Lee, Y., Lee, J., Hwang, S.~J., Yang, E., and Choi, S.
\newblock Neural complexity measures.
\newblock In Larochelle, H., Ranzato, M., Hadsell, R., Balcan, M., and Lin, H.
  (eds.), \emph{Advances in Neural Information Processing Systems 33: Annual
  Conference on Neural Information Processing Systems 2020, NeurIPS 2020,
  December 6-12, 2020, virtual}, 2020.
\newblock URL
  \url{https://proceedings.neurips.cc/paper/2020/hash/6e17a5fd135fcaf4b49f2860c2474c7c-Abstract.html}.

\bibitem[Liu et~al.(2019)Liu, Ott, Goyal, Du, Joshi, Chen, Levy, Lewis,
  Zettlemoyer, and Stoyanov]{liu2019roberta}
Liu, Y., Ott, M., Goyal, N., Du, J., Joshi, M., Chen, D., Levy, O., Lewis, M.,
  Zettlemoyer, L., and Stoyanov, V.
\newblock Roberta: {A} robustly optimized {BERT} pretraining approach.
\newblock \emph{CoRR}, abs/1907.11692, 2019.
\newblock URL \url{http://arxiv.org/abs/1907.11692}.

\bibitem[Maas et~al.(2011)Maas, Daly, Pham, Huang, Ng, and Potts]{maas2011imdb}
Maas, A.~L., Daly, R.~E., Pham, P.~T., Huang, D., Ng, A.~Y., and Potts, C.
\newblock Learning word vectors for sentiment analysis.
\newblock In \emph{Proceedings of the 49th Annual Meeting of the Association
  for Computational Linguistics: Human Language Technologies}, pp.\  142--150,
  Portland, Oregon, USA, June 2011. Association for Computational Linguistics.
\newblock URL \url{http://www.aclweb.org/anthology/P11-1015}.

\bibitem[Mahabadi et~al.(2021)Mahabadi, Belinkov, and
  Henderson]{mahabadi2021ib}
Mahabadi, R.~K., Belinkov, Y., and Henderson, J.
\newblock Variational information bottleneck for effective low-resource
  fine-tuning.
\newblock In \emph{9th International Conference on Learning Representations,
  {ICLR} 2021, Virtual Event, Austria, May 3-7, 2021}. OpenReview.net, 2021.
\newblock URL \url{https://openreview.net/forum?id=kvhzKz-\_DMF}.

\bibitem[McCoy et~al.(2019)McCoy, Pavlick, and Linzen]{mccoy-etal-2019-right}
McCoy, T., Pavlick, E., and Linzen, T.
\newblock Right for the wrong reasons: Diagnosing syntactic heuristics in
  natural language inference.
\newblock In \emph{Proceedings of the 57th Annual Meeting of the Association
  for Computational Linguistics}, pp.\  3428--3448, Florence, Italy, July 2019.
  Association for Computational Linguistics.
\newblock \doi{10.18653/v1/P19-1334}.
\newblock URL \url{https://aclanthology.org/P19-1334}.

\bibitem[Naik et~al.(2018)Naik, Ravichander, Sadeh, Ros{\'{e}}, and
  Neubig]{naik2021stress}
Naik, A., Ravichander, A., Sadeh, N.~M., Ros{\'{e}}, C.~P., and Neubig, G.
\newblock Stress test evaluation for natural language inference.
\newblock In Bender, E.~M., Derczynski, L., and Isabelle, P. (eds.),
  \emph{Proceedings of the 27th International Conference on Computational
  Linguistics, {COLING} 2018, Santa Fe, New Mexico, USA, August 20-26, 2018},
  pp.\  2340--2353. Association for Computational Linguistics, 2018.
\newblock URL \url{https://aclanthology.org/C18-1198/}.

\bibitem[Orgad et~al.(2022)Orgad, Goldfarb{-}Tarrant, and
  Belinkov]{orgad2022debiasing}
Orgad, H., Goldfarb{-}Tarrant, S., and Belinkov, Y.
\newblock How gender debiasing affects internal model representations, and why
  it matters.
\newblock \emph{CoRR}, abs/2204.06827, 2022.
\newblock \doi{10.48550/arXiv.2204.06827}.
\newblock URL \url{https://doi.org/10.48550/arXiv.2204.06827}.

\bibitem[Ravichander et~al.(2021)Ravichander, Dalmia, Ryskina, Metze, Hovy, and
  Black]{ravichander-etal-2021-noiseqa}
Ravichander, A., Dalmia, S., Ryskina, M., Metze, F., Hovy, E., and Black, A.~W.
\newblock {N}oise{QA}: Challenge set evaluation for user-centric question
  answering.
\newblock In \emph{Proceedings of the 16th Conference of the European Chapter
  of the Association for Computational Linguistics: Main Volume}, pp.\
  2976--2992, Online, April 2021. Association for Computational Linguistics.
\newblock \doi{10.18653/v1/2021.eacl-main.259}.
\newblock URL \url{https://aclanthology.org/2021.eacl-main.259}.

\bibitem[Rosenfeld et~al.(2020)Rosenfeld, Rosenfeld, Belinkov, and
  Shavit]{rosenfield2020scaling}
Rosenfeld, J.~S., Rosenfeld, A., Belinkov, Y., and Shavit, N.
\newblock A constructive prediction of the generalization error across scales.
\newblock In \emph{8th International Conference on Learning Representations,
  {ICLR} 2020, Addis Ababa, Ethiopia, April 26-30, 2020}. OpenReview.net, 2020.
\newblock URL \url{https://openreview.net/forum?id=ryenvpEKDr}.

\bibitem[Russakovsky et~al.(2015)Russakovsky, Deng, Su, Krause, Satheesh, Ma,
  Huang, Karpathy, Khosla, Bernstein, Berg, and
  Fei-Fei]{russakovsky2015imagenet}
Russakovsky, O., Deng, J., Su, H., Krause, J., Satheesh, S., Ma, S., Huang, Z.,
  Karpathy, A., Khosla, A., Bernstein, M., Berg, A.~C., and Fei-Fei, L.
\newblock {ImageNet Large Scale Visual Recognition Challenge}.
\newblock \emph{International Journal of Computer Vision (IJCV)}, 115\penalty0
  (3):\penalty0 211--252, 2015.
\newblock \doi{10.1007/s11263-015-0816-y}.

\bibitem[Sanh et~al.(2019)Sanh, Debut, Chaumond, and Wolf]{sanh2019distilbert}
Sanh, V., Debut, L., Chaumond, J., and Wolf, T.
\newblock Distilbert, a distilled version of {BERT:} smaller, faster, cheaper
  and lighter.
\newblock \emph{CoRR}, abs/1910.01108, 2019.
\newblock URL \url{http://arxiv.org/abs/1910.01108}.

\bibitem[Saphra \& Lopez(2019)Saphra and
  Lopez]{saphra-lopez-2019-understanding}
Saphra, N. and Lopez, A.
\newblock Understanding learning dynamics of language models with {SVCCA}.
\newblock In \emph{Proceedings of the 2019 Conference of the North {A}merican
  Chapter of the Association for Computational Linguistics: Human Language
  Technologies, Volume 1 (Long and Short Papers)}, pp.\  3257--3267,
  Minneapolis, Minnesota, June 2019. Association for Computational Linguistics.
\newblock \doi{10.18653/v1/N19-1329}.
\newblock URL \url{https://aclanthology.org/N19-1329}.

\bibitem[Saxe et~al.(2018)Saxe, Bansal, Dapello, Advani, Kolchinsky, Tracey,
  and Cox]{saxe2018ib}
Saxe, A.~M., Bansal, Y., Dapello, J., Advani, M., Kolchinsky, A., Tracey,
  B.~D., and Cox, D.~D.
\newblock On the information bottleneck theory of deep learning.
\newblock In \emph{6th International Conference on Learning Representations,
  {ICLR} 2018, Vancouver, BC, Canada, April 30 - May 3, 2018, Conference Track
  Proceedings}. OpenReview.net, 2018.
\newblock URL \url{https://openreview.net/forum?id=ry\_WPG-A-}.

\bibitem[Sch{\"{o}}lkopf et~al.(2012)Sch{\"{o}}lkopf, Janzing, Peters,
  Sgouritsa, Zhang, and Mooij]{scholkopf2012causal}
Sch{\"{o}}lkopf, B., Janzing, D., Peters, J., Sgouritsa, E., Zhang, K., and
  Mooij, J.~M.
\newblock On causal and anticausal learning.
\newblock In \emph{Proceedings of the 29th International Conference on Machine
  Learning, {ICML} 2012, Edinburgh, Scotland, UK, June 26 - July 1, 2012}.
  icml.cc / Omnipress, 2012.
\newblock URL \url{http://icml.cc/2012/papers/625.pdf}.

\bibitem[Shen et~al.(2021)Shen, Liu, He, Zhang, Xu, Yu, and Cui]{shen2021ood}
Shen, Z., Liu, J., He, Y., Zhang, X., Xu, R., Yu, H., and Cui, P.
\newblock Towards out-of-distribution generalization: {A} survey.
\newblock \emph{CoRR}, abs/2108.13624, 2021.
\newblock URL \url{https://arxiv.org/abs/2108.13624}.

\bibitem[Shwartz{-}Ziv \& Tishby(2017)Shwartz{-}Ziv and Tishby]{ziv2017ib}
Shwartz{-}Ziv, R. and Tishby, N.
\newblock Opening the black box of deep neural networks via information.
\newblock \emph{CoRR}, abs/1703.00810, 2017.
\newblock URL \url{http://arxiv.org/abs/1703.00810}.

\bibitem[Swayamdipta et~al.(2020)Swayamdipta, Schwartz, Lourie, Wang,
  Hajishirzi, Smith, and Choi]{swayamdipta2020cartography}
Swayamdipta, S., Schwartz, R., Lourie, N., Wang, Y., Hajishirzi, H., Smith,
  N.~A., and Choi, Y.
\newblock Dataset cartography: Mapping and diagnosing datasets with training
  dynamics.
\newblock In Webber, B., Cohn, T., He, Y., and Liu, Y. (eds.),
  \emph{Proceedings of the 2020 Conference on Empirical Methods in Natural
  Language Processing, {EMNLP} 2020, Online, November 16-20, 2020}, pp.\
  9275--9293. Association for Computational Linguistics, 2020.
\newblock \doi{10.18653/v1/2020.emnlp-main.746}.
\newblock URL \url{https://doi.org/10.18653/v1/2020.emnlp-main.746}.

\bibitem[T{\"{a}}nzer et~al.(2021)T{\"{a}}nzer, Ruder, and
  Rei]{tanzer2021bertmem}
T{\"{a}}nzer, M., Ruder, S., and Rei, M.
\newblock {BERT} memorisation and pitfalls in low-resource scenarios.
\newblock \emph{CoRR}, abs/2105.00828, 2021.
\newblock URL \url{https://arxiv.org/abs/2105.00828}.

\bibitem[Tishby \& Zaslavsky(2015)Tishby and Zaslavsky]{tishby2015ib}
Tishby, N. and Zaslavsky, N.
\newblock Deep learning and the information bottleneck principle.
\newblock In \emph{2015 {IEEE} Information Theory Workshop, {ITW} 2015,
  Jerusalem, Israel, April 26 - May 1, 2015}, pp.\  1--5. {IEEE}, 2015.
\newblock \doi{10.1109/ITW.2015.7133169}.
\newblock URL \url{https://doi.org/10.1109/ITW.2015.7133169}.

\bibitem[Tishby et~al.(1999)Tishby, Pereira, and Bialek]{tishby99information}
Tishby, N., Pereira, F.~C., and Bialek, W.
\newblock The information bottleneck method.
\newblock In \emph{Proc. of the 37-th Annual Allerton Conference on
  Communication, Control and Computing}, pp.\  368--377, 1999.
\newblock URL \url{https://arxiv.org/abs/physics/0004057}.

\bibitem[Tu et~al.(2020)Tu, Lalwani, Gella, and He]{tu2020bert}
Tu, L., Lalwani, G., Gella, S., and He, H.
\newblock {An Empirical Study on Robustness to Spurious Correlations using
  Pre-trained Language Models}.
\newblock \emph{Transactions of the Association for Computational Linguistics},
  8:\penalty0 621--633, 10 2020.
\newblock ISSN 2307-387X.
\newblock \doi{10.1162/tacl_a_00335}.
\newblock URL \url{https://doi.org/10.1162/tacl\_a\_00335}.

\bibitem[Voita \& Titov(2020)Voita and Titov]{voita2020mdl}
Voita, E. and Titov, I.
\newblock Information-theoretic probing with minimum description length.
\newblock In Webber, B., Cohn, T., He, Y., and Liu, Y. (eds.),
  \emph{Proceedings of the 2020 Conference on Empirical Methods in Natural
  Language Processing, {EMNLP} 2020, Online, November 16-20, 2020}, pp.\
  183--196. Association for Computational Linguistics, 2020.
\newblock \doi{10.18653/v1/2020.emnlp-main.14}.
\newblock URL \url{https://doi.org/10.18653/v1/2020.emnlp-main.14}.

\bibitem[Wang et~al.(2021{\natexlab{a}})Wang, Wang, Cheng, Gan, Jia, Li, and
  Liu]{wang2021infobert}
Wang, B., Wang, S., Cheng, Y., Gan, Z., Jia, R., Li, B., and Liu, J.
\newblock Infobert: Improving robustness of language models from an information
  theoretic perspective.
\newblock In \emph{9th International Conference on Learning Representations,
  {ICLR} 2021, Virtual Event, Austria, May 3-7, 2021}. OpenReview.net,
  2021{\natexlab{a}}.
\newblock URL \url{https://openreview.net/forum?id=hpH98mK5Puk}.

\bibitem[Wang et~al.(2021{\natexlab{b}})Wang, Lan, Liu, Ouyang, and
  Qin]{wang2021ood}
Wang, J., Lan, C., Liu, C., Ouyang, Y., and Qin, T.
\newblock Generalizing to unseen domains: {A} survey on domain generalization.
\newblock In Zhou, Z. (ed.), \emph{Proceedings of the Thirtieth International
  Joint Conference on Artificial Intelligence, {IJCAI} 2021, Virtual Event /
  Montreal, Canada, 19-27 August 2021}, pp.\  4627--4635. ijcai.org,
  2021{\natexlab{b}}.
\newblock \doi{10.24963/ijcai.2021/628}.
\newblock URL \url{https://doi.org/10.24963/ijcai.2021/628}.

\bibitem[Zhang et~al.(2017)Zhang, Bengio, Hardt, Recht, and
  Vinyals]{zhang_understanding_2017}
Zhang, C., Bengio, S., Hardt, M., Recht, B., and Vinyals, O.
\newblock Understanding deep learning requires rethinking generalization.
\newblock \emph{arXiv:1611.03530 [cs]}, February 2017.
\newblock URL \url{http://arxiv.org/abs/1611.03530}.
\newblock arXiv: 1611.03530.

\bibitem[Zhang et~al.(2022)Zhang, He, Xu, Yu, Shen, and Cui]{zhang2022nico}
Zhang, X., He, Y., Xu, R., Yu, H., Shen, Z., and Cui, P.
\newblock Nico++: Towards better benchmarking for domain generalization, 2022.
\newblock URL \url{https://arxiv.org/abs/2204.08040}.

\bibitem[Zhao et~al.(2018)Zhao, Dua, and Singh]{zhao2018generating}
Zhao, Z., Dua, D., and Singh, S.
\newblock Generating natural adversarial examples.
\newblock In \emph{6th International Conference on Learning Representations,
  {ICLR} 2018, Vancouver, BC, Canada, April 30 - May 3, 2018, Conference Track
  Proceedings}. OpenReview.net, 2018.
\newblock URL \url{https://openreview.net/forum?id=H1BLjgZCb}.

\end{thebibliography}
\bibliographystyle{references}

\newpage

\appendix
\section{Computing Information Measures}
\label{sec:computing}

Our goal is to compute the Entropy for a neuron, $A_x$, and 
MI between a pair of neurons, $A_x$ and $A_y$, 
for some $(x, y) \in \{1,\dots,N\}$,
where the neurons are represented 
as their activations for a given number of samples (S). 

\subsection{Entropy}
\label{subsec:computing_entropy}

\begin{algorithm}
\caption{Computing entropy}\label{alg:entropy_computation}
\begin{algorithmic}[1]
\Procedure{Entropy}{$A_x$}
    \State $H(A_x) \gets 0$
    \State $\Hat{A}_{x} \gets \textsc{Bin}(A{(x,i)})$
    \Comment{Discretizing activations by binning}
    \For{$i \in \{1,\dots,N_{bins}\}$}
        \State $p(\Hat{a}_{(x,i)}) \gets \{\mathbbm{1}\{\Hat{a}_{(x,i)} == \Hat{a}_{(x,j)}\}\}_{j=1}^{S}/S$
        \Comment{Computing probability}
        \State $h(a_{(x,i)}) \gets p(\Hat{a}_{(x,i)})\log(1/p(\Hat{a}_{(x,i)}))$
        \Comment{As per Eq. \ref{eq:entropy}}
        \State $H(A_x) \gets H(A_x) + h(a_{(x,i)})$
    \EndFor
    \State $H(A_x) \gets H(A_x)/S$
    \Comment{As per Eq. \ref{eq:entropy}}
    \State \textbf{return} $H(A_x)$
\EndProcedure
\end{algorithmic}
\end{algorithm}

In order to compute entropy, 
we first discretize the set of neuron activations
by binning them in a uniform range of values
\citep{darbellay1999info}.
Specifically, we divide the range of activations for a neurons
into a constant number of bins (usually $100$),
each of the same size.
Then, the probability for each activation is determined
by the number of activations that share the same discrete value.
Plugging these values in Equation \ref{eq:entropy},
across all activations for a neurons
gives us the entropy for that neuron.

\subsection{Mutual Information}
\label{subsec:computing_mi}
%
To compute the MI between these one-dimensional variables, we use the estimator proposed in \cite{kraskov2004mi}, which provides a tight lower bound to the mutual information, especially in low-dimensional cases~\citep{belghazi2018mi}.

\cite{kraskov2004mi} consider the popular interpretation of mutual information $I(X;Y)$ 
between two continuous random variables $X$ and $Y$ as ``the amount of uncertainty left in $Y$ when $X$ is known". In terms of Shannon entropy, this is equivalent to $I(X;Y)=H(Y)-H(Y|X) \equiv H(X) + H(Y) - H(X, Y)$.
Following 
this formalization,
we require to work with three variables- $A_x$, $A_y$, and $A_z = (A_x, A_y)$. 
To compute the MI, the $k$-nearest neighbor distances are considered for these variables. 
Particularly, for some given distance metric, we represent $\epsilon_i$ as the distance of a given point $a_{(z, i)} = (a_{(x, i)}, a_{(y, i)})$ to its $k^{th}$-neighbor. This distance is then considered in the $x$ and $y$ spaces---$e_{(x, i)}$ and $e_{(y, i)}$ depict the number of points that lie at a distance lesser than $\epsilon_{i}$ with respect to $a_{(x, i)}$ and $a_{(y, i)}$, respectively.
Then, the mutual information is given as:

\begin{equation}
    I(A_x; A_y) = \psi(k) + \psi(S) - \frac{1}{S}\sum^{S}_{i=1}(\psi(e_{(x, i)}) + \psi(e_{(y, i)}))
    \label{eq:kraskov}
\end{equation}
where, $\psi(\cdot)$ is the digamma function, or the logarithmic derivative of the gamma function $\Gamma(\cdot)$: 
$\psi(x) = \frac{\Gamma'(x)}{\Gamma(x)} = \ln{x} - \frac{1}{2x}$.

We use the Chebyshev distance or the $L_{\infty}$ metric as our distance metric for all distance computations in the $X$, $Y$, and $Z$ spaces.

\begin{algorithm}
\caption{Computing mutual information}
\label{alg:mi_computation}
\begin{algorithmic}[1]
\Procedure{MI}{$A_x$, $A_y$}
    \State $I(A_x; A_y) \gets \psi(k) + \psi(S)$
    \Comment{As per Eq. \ref{eq:kraskov}}
    \State $A_z \gets A_x \bigoplus A_y$
    \State $\textsc{KNN} \gets \textsc{KNN}(A_z)$
    \Comment{Initializing k-nearest neighbor distances}
    \State $e_{x} \gets 0$, $e_{y} \gets 0$
    \For{$i \in \{1,\dots,S\}$}
        \State $\epsilon_{i} \gets \textsc{KNN}(a_{(z, i)}).r(k)$
        \Comment{Distance to the $k^{th}$ neighbor}
        \State $e_{(x, i)} \gets \{\mathbbm{1}\{||x_j-x_i|| < \epsilon_{i}\}\}_{j=1}^{X-i}$
        \State $e_{(y, i)} \gets \{\mathbbm{1}\{||y_j-y_i|| < \epsilon_{i}\}\}_{j=1}^{X-i}$
        \State $e_{x} \gets e_{x} + \psi(e_{(x, i)})$, $e_{y} \gets e_{y} + \psi(e_{(y, i)})$
    \EndFor
    \State $I(A_x; A_y) \gets I(A_x; A_y) - \frac{1}{|X|}(e_{x} + e_{y})$
    \Comment{As per Eq. \ref{eq:kraskov}}
    \State \textbf{return} $I(A_x; A_y)$
\EndProcedure
\end{algorithmic}
\end{algorithm}




Popular recent estimators take a neural approach for estimating MI and consider an alternate interpretation of MI as the dependence between two random variables, i.e., $I(X;Y) = D_{KL}(P_{(X,Y)} || P_X \otimes P_Y) = D_{KL}(P_{(Y|X)} || P_Y)$. 
Since $p(y|x)$ is a-priori unknown for most real-world distributions, \cite{cheng2020club} approximate it through a parametric variational distribution $q_{\theta}(y|x)$. 
\cite{belghazi2018mi} instead consider representations that estimate the KL divergence. They parameterize the family of functions that defines the bound of these representations using a neural network.
However, these approaches require us to learn a distinct neural network for each pair of random variable across which we wish to compute the MI. 
Considering the large number of pairwise operations, especially when working at the level of individual neurons, these approaches are not feasible for our analysis.

\section{Extended Experimental Details}
\label{sec:experimental_details}

\begin{table}[ht]
\centering
\caption{
Values for network and training hyper-parameters
across different experimental settings.
\label{tb:network_parameters}
}
\resizebox{\textwidth}{!}{
\begin{tabular}{l|rrrrrc}
\toprule
\textbf{Model}       & \multicolumn{1}{l}{\textbf{\# Layers}} & \multicolumn{1}{l}{\textbf{\# Parameters}} & \multicolumn{1}{l}{\textbf{\# Epochs}} & \multicolumn{1}{l}{\textbf{LR}} & \multicolumn{1}{l}{\textbf{Batch size}} & \multicolumn{1}{l}{\textbf{Pretrained?}} \\ \midrule
MLP (\memiiSynthSetI)  & 3                                  & 0.232M                                  & 50                                  & 0.001                           & 512                                     & $\times$                                        \\
MLP (\texttt{Shuffled MNIST}) & 3                                  & 0.184M                                  & 200                                 & 0.001                           & 512                                     & $\times$                                        \\
DistilBERT           & 6                                  & 66M                                     & 20                                  & 5e-5                            & 64                                      & \Checkmark                                        \\
RoBERTa-base         & 12                                 & 123M                                    & 10                                  & 5e-5                            & 64                                      & \Checkmark                                       \\
ResNet-18            & 18                                 & 11M                                     & 25                                  & 1e-5                            & 64                                      & $\times$                                        \\ \bottomrule
\end{tabular}}
\end{table}

Details for network architectures and training regimes
are given in Table \ref{tb:network_parameters}.
Further explanations regarding datasets for the various
experimental settings are elaborated below.

\subsection{Colored MNIST}
\label{subsec:details_colored_mnist}
All images in the training and evaluation sets
are colored green or red.
$\alpha$ $\in \{0.00, 0.25, 0.50, 0.75, 1.00\}$
corresponds to the fraction of training examples
that follow the color-to-label correlation:
green $\rightarrow$ label-$0$ (digits $0$--$4$) and
red $\rightarrow$ label-$1$ (digits $5$--$9$).
The size of the set is same as the original MNIST set,
that is a training set of 50k examples and an
evaluation set of 10k examples.
The hyper-parameters for model training
were picked from~\citep{arjovsky2019irm}.

\subsection{Sentiment Adjectives}
\label{subsec:details_adjectives}
The chosen set of adjectives for sub-sampling
examples from \texttt{IMDb} and then induce the adjective-based
spurious correlations are:
\textbf{Positive}: \{\textit{good}, \textit{great}, \textit{wonderful}, \textit{excellent}, \textit{best}\},
and
\textbf{Negative}: \{\textit{bad}, \textit{terrible}, \textit{awful}, \textit{poor}, \textit{negative}\}.
%
%
To maintain the correlation for each individual adjective,
the set does not contains examples 
that contain adjectives from both of the sets.
The size of this sampled set came to be $\sim$6k,
out of which $67\%$ of the example already abide
to the heuristic. That is, the true label corresponding
to these examples is same as the nature of adjective(s) 
they contain. Thus, we vary the labels corresponding
to the other $33\%$ of the examples with increasing
$\alpha$. Particularly, $\alpha$ takes a value in
$\{0.00, 0.25, 0.50, 0.75, 1.00\}$, where
$0.00$ suggests that $33\%$ of the data does not
follow the heuristic, while $1.00$ suggests that
all examples follow the heuristic.
The learning rate (LR) and batch size was chosen after a 
hyper-parameter sweep with values 
$\{5e-3, 3e-4, 5e-5\}$, $\{16, 32, 64\}$
on the generalizing training set ($\alpha = \beta = 0.00$).

\subsection{Bias-in-Bios}
\label{subsec:details_bias_bios}
We perform our analysis over trained models released by 
\citet{orgad2022debiasing} for randomly picked random seeds
of 0, 5, 26, 42, and 63. 
\memiiNatSetI~\citep{de-arteaga2019biasinbios}
contains around 400k biographies,
the networks are trained on $65\%$ of this set,
while a sub-sampled balancing from the rest of examples
are used for evaluation and analysis.

\subsection{NICO++}
\label{subsec:details_nico}
We sub-sample $6$ animal classes with the most number of examples across them
(\textit{bear}, \textit{dog}, \textit{cat}, \textit{bird}, \textit{horse}, \textit{sheep})
from the original dataset, spanning $\sim$10k training and $\sim$7.8k evaluation examples.
Both \texttt{balanced} and \texttt{unbalanced} set contain all individual contexts
for the animals.
For the \texttt{unbalanced} set:
one from ten common contexts is associated to each of the chosen animal classes
(\textit{grass}, \textit{water}, \textit{autumn}, \textit{dim}, \textit{outdoor}, \textit{rock},
respectively). On the other hand, the \texttt{balanced} set contains equal number of images from all these
contexts for each of the animal classes.

\subsection{Shuffled MNIST}
\label{subsec:details_shuffled_mnist}
Labels are shuffled for the $10$ digits of MNIST
over the 50k training examples.
$\beta$ $\in \{0.00, 0.25, 0.50, 0.75, 1.00\}$.
The evaluation and analysis is performed over 10k
balanced testing examples from the original set.

\subsection{Shuffled IMDb}
\label{subsec:details_imdb}
The networks are trained for the 25k training examples
and shuffled for $\beta$ $\in \{0.00, 0.25, 0.50, 0.75, 1.00\}$.
Performance computations and analysis is performed on
5k examples sampled from the evaluation set.


\section{Extended Experimental Results}
\label{sec:extended_results}

\subsection{Semi-synthetic Heuristic Memorization}
\label{subsec:semi_synthetic_memii_all_results}

\begin{figure}[ht]
    \centering
    \includegraphics[width=\textwidth]{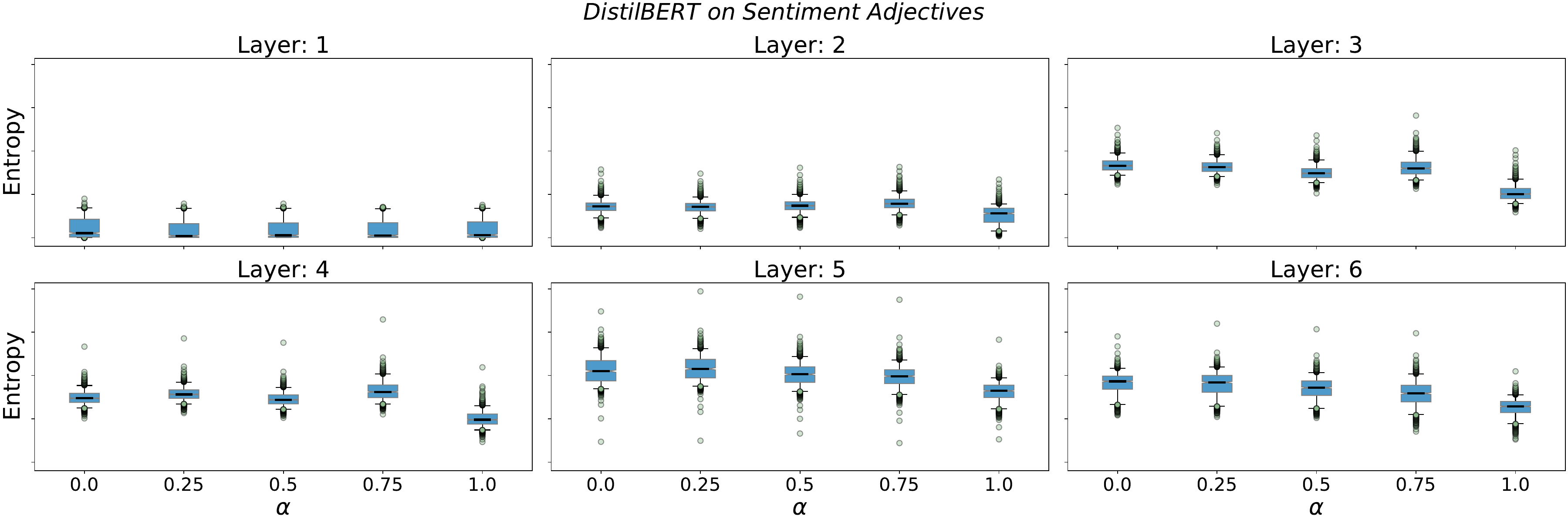}
    \caption{
    Variation of neuron entropy
    across all layers of DistilBERT
    on \memiiSynthSetII.
    \label{fig:sentiment_adj_all_layers}
    }
    \label{fig:sentiment_adj_all_layers_mi}
\end{figure}

Figure \ref{fig:sentiment_adj_all_layers_mi}
shows the variation of neuron entropy across
increasing values of $\alpha$ 
(here, adjective-to-sentiment correlation during training)
for all layers of DistilBERT.
As discussed in the main text (\cref{subsec:semi_synthetic_memii}),
we primarily see a variation in entropy for later layers in the network.
In the first layer, we observe that entropy for all values of $\alpha$
remains similar and low, while as we go towards the later layers,
we start to see certain differences in entropy, with the network for 
$alpha=1.0$ showing especially lower entropy.
Since all these networks have been fine-tuned from the same pre-trained 
initialization and vary only in terms of how much spurious correlation 
is present in their training sets, we attribute this pattern across layers
to a possibility that such spurious correlation are largely captured
in the later layers during fine-tuning.

\subsection{Effect of Model Capacity}
\label{subsec:capacity_effect}

\begin{figure*}[t]
    \centering
    \begin{subfigure}[t]{0.48\linewidth}
        \centering
        \includegraphics[width=\textwidth]{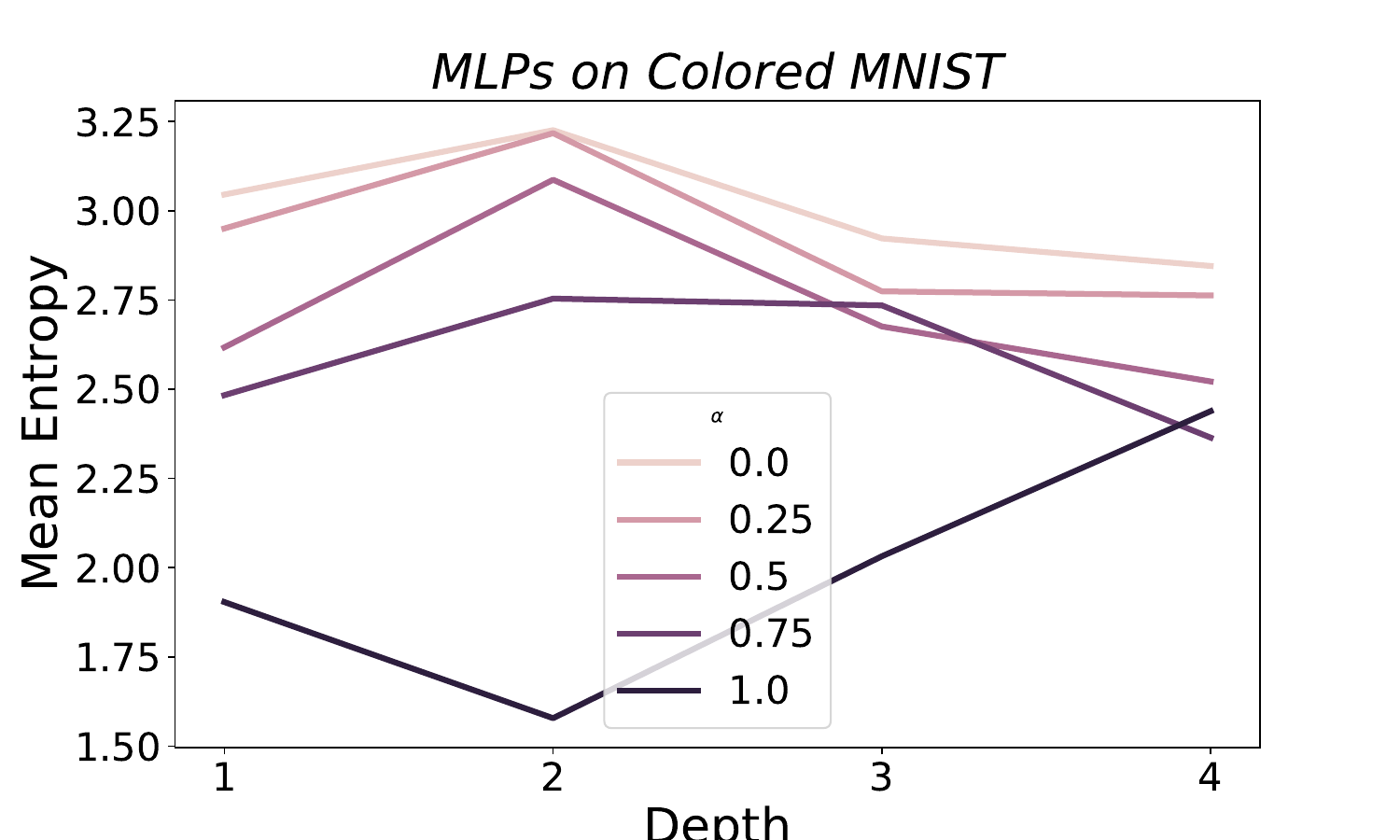}
        \label{sfig:colored_mnist_variation}
    \end{subfigure}
    \hfill
    \begin{subfigure}[t]{0.48\linewidth}
        \centering
        \includegraphics[width=\textwidth]{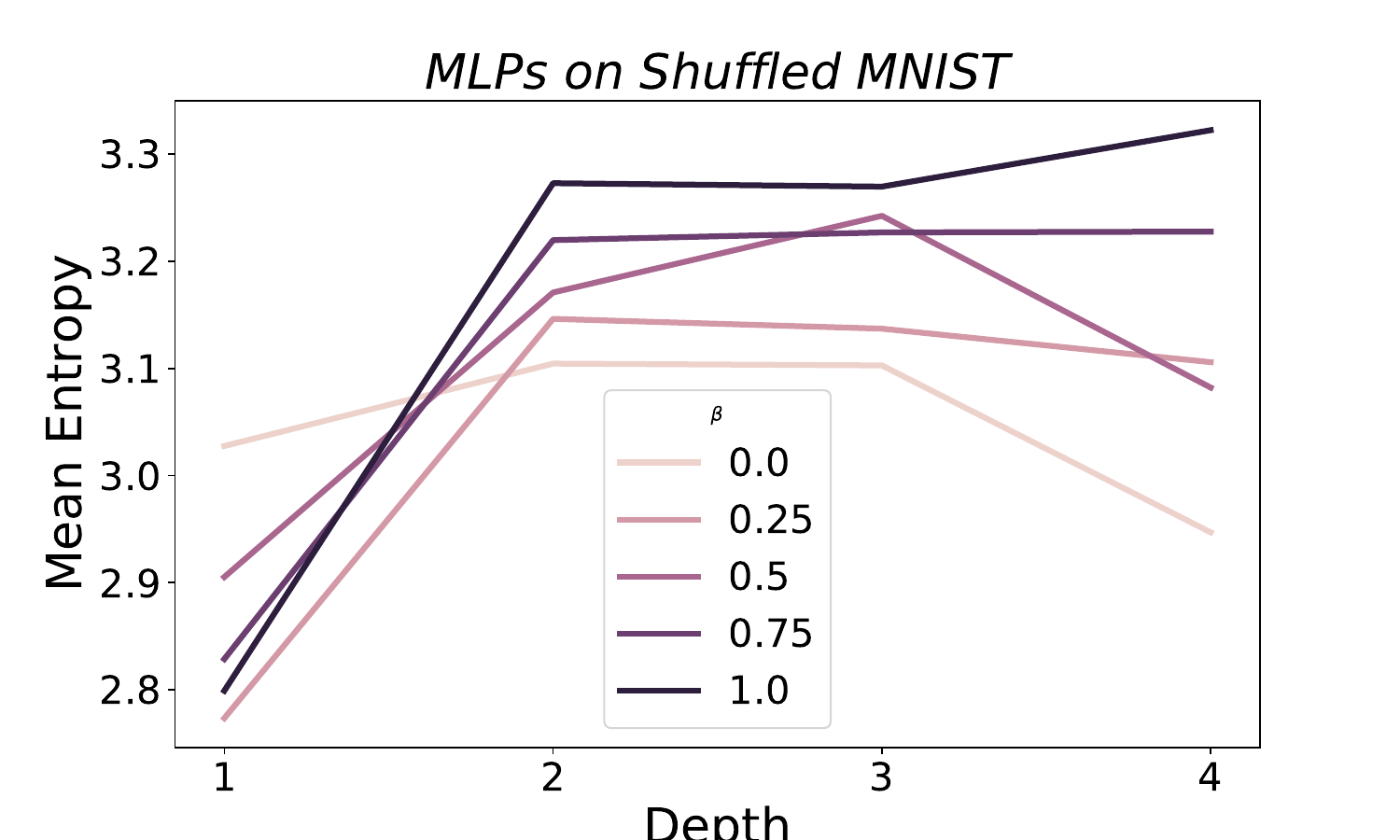}
        \label{sfig:shuffled_mnist_variation}
    \end{subfigure}
    \caption{
    Comparison of entropy for MLP networks
    with varying number of layers. 
    \label{fig:capacity_variation}
    }
\end{figure*}

Theoretically, we expect that capacity of a network would influence entropy and MI. To test this, we train MLP networks with varying depths for the tasks of Colored MNIST and Shuffled MNIST, and compare the neuron entropy for the obtained networks. From results shown in Figure \ref{fig:capacity_variation} observe that our hypothesis holds true for all capacities (ranging from single to four-layered networks). However, different capacity networks show entropy values in different ranges and hence they are likely not directly comparable with each other.

\subsection{Bias-in-Bios variation across layers}
\label{subsec:bias_mi_all_layers}

\begin{figure}[ht]
    \centering
    \includegraphics[width=\textwidth]{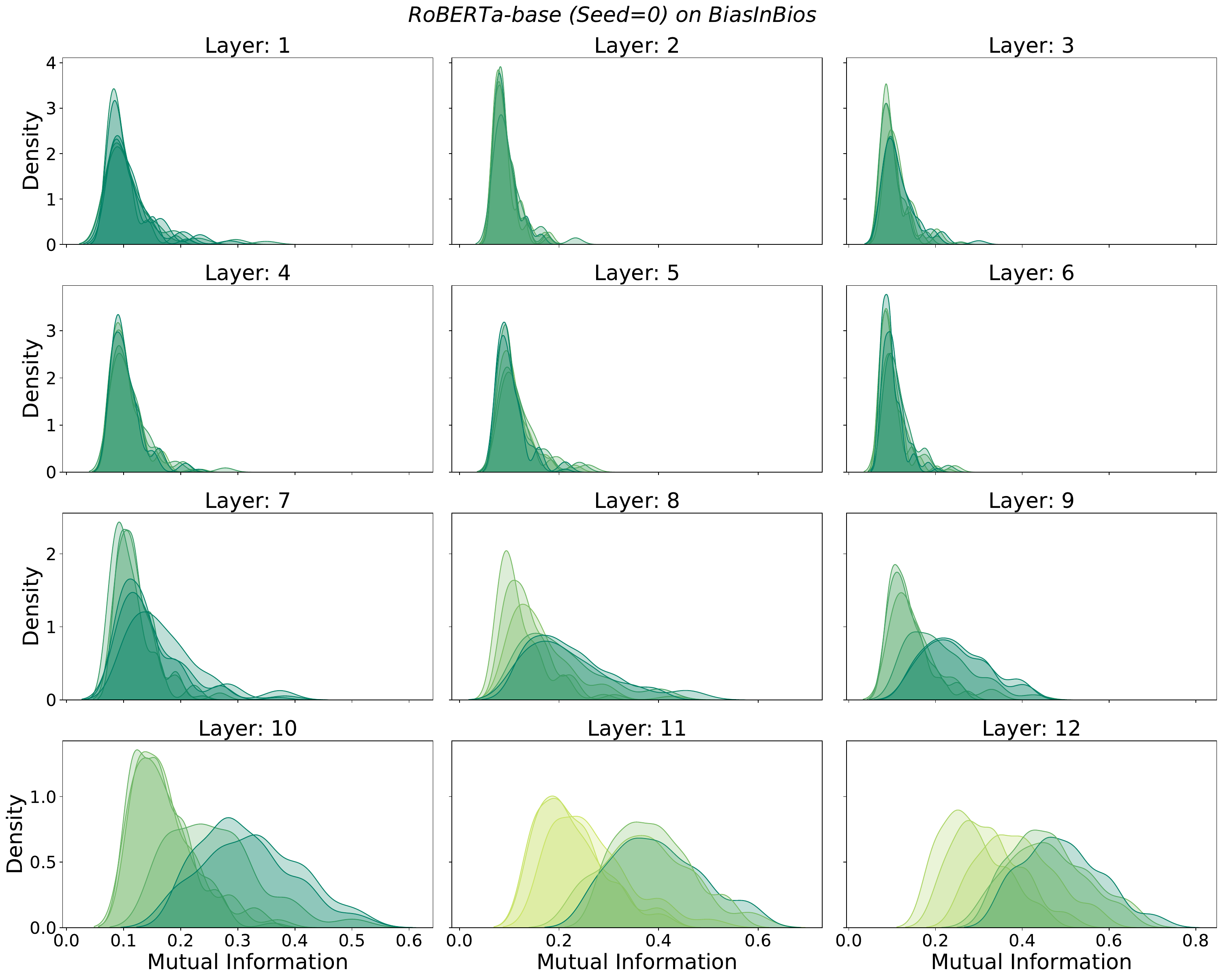}
    \caption{
    Distribution of MI
    across neuron pairs for different layers
    in RoBERTa-base fine-tuned on
    \memiiNatSetI\ training sets.
    Colors represent the value of 
    gender extractability with the MDL probe.
    \label{fig:biasinbios_all_layers_mi}
    }
\end{figure}

Figure \ref{fig:biasinbios_all_layers_mi} shows how the density
of neuron pairs across MI values changes across layers of RoBERTa-base
when trained on different training sets of \memiiNatSetI.
By looking at these distributions of MI in conjunction with
the value of compression (obtained through the MDL probe),
we observe a very interesting phenomenon:
The relation between MI and compression
is maintained throughout the network across its different layers.
Networks trained on the different training sets have very similar
presence of gender information, as indicated by compression values
(color of the Gaussian plots).
This is replicated in the MI distribution where all plots are
clustered at very similar values.
We see that the distinction among the different sets stars to become
more lucid with later layers (reflected in the colors)
and the same distinction is observed with MI, where
more biased models (darker colors) show higher MI.
We reported the results for the last layer 
in the main text (\cref{subsec:natural_memii})
because compression values vary the most at this layer,
and is consequently reflected in MI.
However, looking at the variation across all layers
here validates the usage of information measures
as a tool for evaluation even strongly.

\subsection{Relation to Complexity Measures}
\label{subsec:complexity_measures}
\begin{table}[h]
\centering
\caption{
Pearson correlation coefficient (PCC)
between norm-based complexity measures
and information measures for
the two memorization types.
\label{tb:complexity_pcc}
}
\begin{tabular}{c|cc|cc}
\toprule
                                                                                & \multicolumn{2}{c|}{PCC for Colored MNIST}                  & \multicolumn{2}{c}{PCC for Shuffled MNIST}                  \\ \cmidrule{2-5}
\multirow{-2}{*}{\begin{tabular}[c]{@{}c@{}}Complexity\\ Measures\end{tabular}} & \textbf{Mean Entropy}        & \textbf{Mean MI}             & \textbf{Mean Entropy}        & \textbf{Mean MI}             \\ \hline
2-Norm                                                                          & \cellcolor[HTML]{E0E0E0}0.31 & \cellcolor[HTML]{F1F1F1}0.14 & \cellcolor[HTML]{9E9E9E}0.96 & \cellcolor[HTML]{999999}1    \\
Frobenius-Norm                                                                  & \cellcolor[HTML]{CCCCCC}0.5  & \cellcolor[HTML]{E0E0E0}0.31 & \cellcolor[HTML]{9B9B9B}0.99 & \cellcolor[HTML]{9F9F9F}0.95 \\
Path-Norm                                                                       & \cellcolor[HTML]{C4C4C4}0.58 & \cellcolor[HTML]{E1E1E1}0.3  & \cellcolor[HTML]{9C9C9C}0.98 & \cellcolor[HTML]{A0A0A0}0.94 \\
Validation Acc.                                                                 & \cellcolor[HTML]{9E9E9E}0.96 & \cellcolor[HTML]{A9A9A9}0.85 & \cellcolor[HTML]{A7A7A7}0.87 & \cellcolor[HTML]{B8B8B8}0.7  \\ \bottomrule
\end{tabular}
\end{table}

Here, we validate that the activation diversity does not
simply capture the complexity of a learned model.
Popular norm-based measures are often used as complexity measures (also used for model selection; see \cref{subsec:model_selection_comparison}). 
On computing the absolute pearson correlation 
between these complexity measures
and our proposed information measures
in Table \ref{tb:complexity_pcc}, 
we observe that the two are weakly correlated
for \memiiSynthSetI\ and strongly correlated
for \texttt{Shuffled} \memiSynthSetI.
This suggests that the information measures might
approximate to model complexity in the case of
example-level memorization, but captures more than just
complexity as evident for heuristic memorization.
%
Notably, in both cases, our measures are strongly correlated to the validation accuracy.

\subsection{Variance of Entropy and MI}
As seen from results on our various experimental setups (Figures 1-7),
typically a wide distribution of values is observed for both entropy and MI,
for all models trained on different $\alpha$ and $\beta$ values.
Since entropy is computed for each neuron in the network and MI for all pairs of neurons, a large number of values are obtained for each network, resulting to 
this wide distribution. 
Even though these distributions have a wide range, an observable difference is seen that follows the expected trend across both kinds of memorizations. This is further apparent through the model selection results where even the mean over these two distributions yields encouraging rankings.
However, such a variance could make it challenging for practitioners to
identify differences in cases when the amount of memorization between two models
is not high (for instance, a difference of $\alpha=0.5$).

\section{Extended Analysis for Model Selection}
\label{sec:extended_model_selection}

\subsection{Model Rankings for Bias-in-Bios}
\label{subsec:biasinbias_model_rankings}

\begin{table}[ht]
\centering
\caption{
Kendall's $\tau$ between model rankings from all pairs of 
intrinsic and extrinsic metrics. 
$\tau$ can range from -1.0 (perfect disagreement) 
to 1.0 (perfect agreement).
\label{tb:all_corrs_biasbios}
}
\begin{tabular}{c|ccccccc}
\toprule
\multicolumn{1}{l|}{}                                    & \multicolumn{1}{c}{\begin{tabular}[c]{@{}c@{}}Comp-\\ ression\end{tabular}} & \multicolumn{1}{c}{\begin{tabular}[c]{@{}c@{}}TPR\\ Gap\end{tabular}} & \multicolumn{1}{c}{\begin{tabular}[c]{@{}c@{}}FPR\\ Gap\end{tabular}} & \multicolumn{1}{c}{\begin{tabular}[c]{@{}c@{}}Separation\\ Gap\end{tabular}} & \multicolumn{1}{c}{\begin{tabular}[c]{@{}c@{}}Sufficiency\\ Gap\end{tabular}} & \multicolumn{1}{c}{\begin{tabular}[c]{@{}c@{}}Entropy\\ (Mean)\end{tabular}} & \multicolumn{1}{c}{\begin{tabular}[c]{@{}c@{}}MI\\ (Mean)\end{tabular}} \\ \midrule
Compression                                              & \cellcolor[HTML]{999999}1.00                                                & \cellcolor[HTML]{D0D0D0}0.47                                          & \cellcolor[HTML]{C2C2C2}0.60                                          & \cellcolor[HTML]{D0D0D0}0.47                                                 & \cellcolor[HTML]{D0D0D0}0.47                                                  & \cellcolor[HTML]{D0D0D0}0.47                                                 & \cellcolor[HTML]{C2C2C2}0.60                                            \\
TPR Gap                                                  & \cellcolor[HTML]{D0D0D0}0.47                                                & \cellcolor[HTML]{999999}1.00                                          & \cellcolor[HTML]{A7A7A7}0.87                                          & \cellcolor[HTML]{B5B5B5}0.73                                                 & \cellcolor[HTML]{B5B5B5}0.73                                                  & \cellcolor[HTML]{EBEBEB}0.20                                                 & \cellcolor[HTML]{F9F9F9}0.07                                            \\
FPR Gap                                                  & \cellcolor[HTML]{C2C2C2}0.60                                                & \cellcolor[HTML]{A7A7A7}0.87                                          & \cellcolor[HTML]{999999}1.00                                          & \cellcolor[HTML]{C2C2C2}0.60                                                 & \cellcolor[HTML]{C2C2C2}0.60                                                  & \cellcolor[HTML]{F9F9F9}0.07                                                 & \cellcolor[HTML]{EBEBEB}0.20                                            \\
Separation Gap                                           & \cellcolor[HTML]{D0D0D0}0.47                                                & \cellcolor[HTML]{B5B5B5}0.73                                          & \cellcolor[HTML]{C2C2C2}0.60                                          & \cellcolor[HTML]{999999}1.00                                                 & \cellcolor[HTML]{999999}1.00                                                  & \cellcolor[HTML]{EBEBEB}0.20                                                 & \cellcolor[HTML]{DDDDDD}0.33                                            \\
Sufficiency Gap                                          & \cellcolor[HTML]{D0D0D0}0.47                                                & \cellcolor[HTML]{B5B5B5}0.73                                          & \cellcolor[HTML]{C2C2C2}0.60                                          & \cellcolor[HTML]{999999}1.00                                                 & \cellcolor[HTML]{999999}1.00                                                  & \cellcolor[HTML]{EBEBEB}0.20                                                 & \cellcolor[HTML]{DDDDDD}0.33                                            \\
\begin{tabular}[c]{@{}c@{}}Entropy (Mean)\end{tabular} & \cellcolor[HTML]{D0D0D0}0.47                                                & \cellcolor[HTML]{EBEBEB}0.20                                          & \cellcolor[HTML]{F9F9F9}0.07                                          & \cellcolor[HTML]{EBEBEB}0.20                                                 & \cellcolor[HTML]{EBEBEB}0.20                                                  & \cellcolor[HTML]{999999}1.00                                                 & \cellcolor[HTML]{C2C2C2}0.60                                            \\
\begin{tabular}[c]{@{}c@{}}MI (Mean)\end{tabular}      & \cellcolor[HTML]{C2C2C2}0.60                                                & \cellcolor[HTML]{F9F9F9}0.07                                          & \cellcolor[HTML]{EBEBEB}0.20                                          & \cellcolor[HTML]{DDDDDD}0.33                                                 & \cellcolor[HTML]{DDDDDD}0.33                                                  & \cellcolor[HTML]{C2C2C2}0.60                                                 & \cellcolor[HTML]{999999}1.00                                            \\ \bottomrule
\end{tabular}
\end{table}

Here, we compute the Kendall $\tau$ ranking correlations between all
ranking metrics for the \memiiNatSetI\ dataset (Table \ref{tb:all_corrs_biasbios}).
We observe that extrinsic metrics 
(TPR, FPR, Separation, and Sufficiency Gap)
share higher correlations between
each other than those with MI and Entropy, that are only weakly positive.
All these extrinsic metrics are computed by training and assessing the networks
on specifically curated sets that are labeled for gender information
and are thus expected to be more similar to each other.
However, we see that compression
(an intrinsic metric that still requires labeled data)
correlates in a similar manner to extrinsic as well as
the information theoretic measures (not requiring labeled data).

\subsection{Comparison with Baseline Generalization Measures}
\label{subsec:model_selection_comparison}

We perform model selection 
on several benchmark generalization measures. 
Norm-based measures are popularly used in literature 
as a measure of generalization~\citep{yiding2019fantastic}, 
and we conduct comparisons with three of them: 
2-norm, Frobenuis-norm, and Path-norm. 
We do this for all our setups across 
example-level and heuristic memorization 
(Path-norm is not computed on the transformer models, 
since no established way exists to do so). 
On comparing the model selection performance 
with our information measures, 
we observe that while these baselines 
(specifically Frobenius-norm and Path-norm) 
do  reasonably well in ranking models 
that perform example-level memorization, 
they do poorly for heuristic memorization. 
This follows from the fact that these metrics 
are conventionally based and tested 
on the more widely known notion of memorization, 
that of individual examples. Where 2-norm gives $\tau=1$ on Shuffled-IMDb, 
it is totally uncorrelated ($\tau=0.00$) for Sentiment Adjectives. 

At the same time, our information measures 
hold positive $\tau$ values across all models and tasks, 
while the baseline measures are inconsistent, 
varying from extremely low to high values across setups. 
Frobenius-norm shows perfect negative and 
positive correlations with Shuffled IMDb and MNIST, 
respectively, depicting that a high norm 
value could mean anything depending on the dataset and model.

\begin{table}[hbt]
\caption{
Expanded results for model selection with added 
Kendall $\tau$ ranking correlations for norm-based
generalization measures~\citep{yiding2019fantastic}.
}
\begin{tabular}{l|crrrrrr}
\toprule
             & \multicolumn{1}{c|}{\cellcolor[HTML]{FAEDE0}\begin{tabular}[c]{@{}c@{}}Sentiment\\ Adjectives\end{tabular}} & \multicolumn{1}{c|}{\cellcolor[HTML]{FAEDE0}\begin{tabular}[c]{@{}c@{}}Colored\\ MNIST\end{tabular}} & \multicolumn{3}{c|}{\cellcolor[HTML]{FAEDE0}Bias-in-Bios}                                                                                                                                                                      & \multicolumn{1}{c|}{\cellcolor[HTML]{ECF3E9}\begin{tabular}[c]{@{}c@{}}Shuffled\\ IMDb\end{tabular}} & \multicolumn{1}{c}{\cellcolor[HTML]{ECF3E9}\begin{tabular}[c]{@{}c@{}}Shuffled\\ MNIST\end{tabular}} \\ \cmidrule{2-8}
             & \multicolumn{1}{l|}{\begin{tabular}[c]{@{}l@{}}Validation\\ Accuracy\end{tabular}}                                & \multicolumn{1}{l|}{\begin{tabular}[c]{@{}l@{}}Validation\\ Accuracy\end{tabular}}                         & \multicolumn{1}{l|}{\begin{tabular}[c]{@{}l@{}}Comp-\\ ression\end{tabular}} & \multicolumn{1}{l|}{\begin{tabular}[c]{@{}l@{}}TPR\\ Gap\end{tabular}} & \multicolumn{1}{l|}{\begin{tabular}[c]{@{}l@{}}Suff.\\ Gap\end{tabular}} & \multicolumn{1}{l|}{\begin{tabular}[c]{@{}l@{}}Validation\\ Accuracy\end{tabular}}                         & \multicolumn{1}{l}{\begin{tabular}[c]{@{}l@{}}Validation\\ Accuracy\end{tabular}}                          \\ \midrule
Mean Entropy & \multicolumn{1}{c|}{\cellcolor[HTML]{C6C6C6}0.80}                                                           & \multicolumn{1}{c|}{\cellcolor[HTML]{B7B7B7}1.00}                                                    & \multicolumn{1}{c|}{\cellcolor[HTML]{DEDEDE}0.47}                                                & \multicolumn{1}{c|}{\cellcolor[HTML]{F1F1F1}0.20}                                          & \multicolumn{1}{c|}{\cellcolor[HTML]{F1F1F1}0.20}                        & \multicolumn{1}{c|}{\cellcolor[HTML]{D4D4D4}0.60}                                                    & \multicolumn{1}{c}{\cellcolor[HTML]{B7B7B7}1.00}                                                                         \\ \hline
Mean MI      & \multicolumn{1}{c|}{\cellcolor[HTML]{C6C6C6}0.80}                                                           & \multicolumn{1}{c|}{\cellcolor[HTML]{B7B7B7}1.00}                                                    & \multicolumn{1}{c|}{\cellcolor[HTML]{D4D4D4}0.60}                                                & \multicolumn{1}{c|}{\cellcolor[HTML]{FBFBFB}0.07}                                          & \multicolumn{1}{c|}{\cellcolor[HTML]{E8E8E8}0.33}                        & \multicolumn{1}{c|}{\cellcolor[HTML]{C6C6C6}0.80}                                                    & \multicolumn{1}{c}{\cellcolor[HTML]{B7B7B7}1.00}                                                                         \\ \hline
2-Norm & \multicolumn{1}{c|}{\cellcolor[HTML]{FFFFFF}0.00}                                                           & \multicolumn{1}{c|}{\cellcolor[HTML]{F1F1F1}-0.20}                                                    & \multicolumn{1}{c|}{\cellcolor[HTML]{E8E8E8}0.33}                                                & \multicolumn{1}{c|}{\cellcolor[HTML]{D4D4D4}0.60}                                          & \multicolumn{1}{c|}{\cellcolor[HTML]{C1C1C1}0.87}                        & \multicolumn{1}{c|}{\cellcolor[HTML]{B7B7B7}1.00}                                                    & \multicolumn{1}{c}{\cellcolor[HTML]{C6C6C6}0.80}                                                                         \\ \hline
Frobenius-Norm & \multicolumn{1}{c|}{\cellcolor[HTML]{DEDEDE}-0.40}                                                           & \multicolumn{1}{c|}{\cellcolor[HTML]{C6C6C6}0.80}                                                    & \multicolumn{1}{c|}{\cellcolor[HTML]{F1F1F1}-0.20}                                                & \multicolumn{1}{c|}{\cellcolor[HTML]{FBFBFB}0.07}                                          & \multicolumn{1}{c|}{\cellcolor[HTML]{FBFBFB}0.07}                        & \multicolumn{1}{c|}{\cellcolor[HTML]{B7B7B7}-1.00}                                                    & \multicolumn{1}{c}{\cellcolor[HTML]{B7B7B7}1.00}                                                                         \\ \hline
Path-Norm & \multicolumn{1}{c|}{\cellcolor[HTML]{FFFFFF}}                                                           & \multicolumn{1}{c|}{\cellcolor[HTML]{C6C6C6}0.80}                                                    & \multicolumn{1}{c|}{\cellcolor[HTML]{FFFFFF}}                                                & \multicolumn{1}{c|}{\cellcolor[HTML]{FFFFFF}}                                          & \multicolumn{1}{c|}{\cellcolor[HTML]{FFFFFF}}                        & \multicolumn{1}{c|}{\cellcolor[HTML]{FFFFFF}}                                                    & \multicolumn{1}{c}{\cellcolor[HTML]{B7B7B7}1.00}                                                                         \\ \hline
\end{tabular}
\end{table}


\section{Compute Resources and Details}
\label{sec:compute_resources}

\begin{table}[ht]
\centering
\caption{
GPU hours used per experiment.
\label{tb:gpu_hours}
}
\resizebox{\textwidth}{!}{
\begin{tabular}{lrrrrrrr}
\toprule
\multicolumn{1}{l|}{}                                                                             & \multicolumn{1}{c}{\begin{tabular}[c]{@{}c@{}}\ \textbf{\# Types}\\ (Dataset/Training\\ Configurations)\end{tabular}} & \multicolumn{1}{c}{\begin{tabular}[c]{@{}c@{}}\ \textbf{\# Seeds}/\\ \textbf{Type}\end{tabular}} & \multicolumn{2}{|c}{\textbf{Training}}                                                                                                                                 & \multicolumn{2}{|c|}{\textbf{Inference}}                                                                                                                                & \multicolumn{1}{c}{\multirow{2}{*}{\begin{tabular}[c]{@{}c@{}}\textbf{Total Time}\\ (Hours)\end{tabular}}} \\ \cline{4-7}
\multicolumn{1}{l|}{}                                                                             & \multicolumn{1}{l}{}                                                                                       & \multicolumn{1}{c}{}                                                         & \multicolumn{1}{|c}{\begin{tabular}[|c|]{@{}c@{}}Time/Run\\ (Hours)\end{tabular}} & \multicolumn{1}{|c}{\begin{tabular}[c]{@{}c@{}}Total\\ (Hours)\end{tabular}} & \multicolumn{1}{|c}{\begin{tabular}[c]{@{}c@{}}Time/Run\\ (Hours)\end{tabular}} & \multicolumn{1}{|c|}{\begin{tabular}[c]{@{}c@{}}Total\\ (Hours)\end{tabular}} & \multicolumn{1}{c}{}                                                                              \\ \midrule
\multicolumn{1}{l|}{MLP (\memiiSynthSetI)}                                                          & 5                                                                                                          & 10                                                                           & 0.25                                                                           & 12.5                                                                        & 0.05                                                                           & 2.5                                                                         & 15                                                                                                \\
\multicolumn{1}{l|}{MLP (\texttt{Shuffled MNIST})}                                                         & 5                                                                                                          & 10                                                                           & 1                                                                              & 50                                                                          & 0.05                                                                           & 2.5                                                                         & 52.5                                                                                              \\
\multicolumn{1}{l|}{\begin{tabular}[c]{@{}l@{}}DistilBERT \\ (\memiiSynthSetII)\end{tabular}} & 5                                                                                                          & 5                                                                            & 1                                                                              & 25                                                                          & 0.20                                                                           & 5                                                                           & 30                                                                                                \\
\multicolumn{1}{l|}{\begin{tabular}[c]{@{}l@{}}DistilBERT\\ (\texttt{Shuffled IMDb})\end{tabular}}         & 5                                                                                                          & 5                                                                            & 2.5                                                                            & 62.5                                                                        & 0.20                                                                           & 5                                                                           & 67.5                                                                                              \\
\multicolumn{1}{l|}{RoBERTa-base}                                                                 & 5                                                                                                          & 5                                                                            & -                                                                              & -                                                                           & 0.5                                                                            & 12.5                                                                        & 12.5                                                                                              \\
\multicolumn{1}{l|}{ResNet-18}                                                                    & 2                                                                                                          & 5                                                                            & 1                                                                              & 10                                                                          & 0.25                                                                           & 2.5                                                                         & 12.5                                                                                              \\ \hline
\textbf{Total GPU Time}                                                                           & \multicolumn{1}{l}{}                                                                                       & \multicolumn{1}{l}{}                                                         & \multicolumn{1}{l}{}                                                           & \multicolumn{1}{l}{}                                                        & \multicolumn{1}{l}{}                                                           & \multicolumn{1}{l}{}                                                        & \textbf{190}                                                                                      \\ \bottomrule
\end{tabular}}
\end{table}

For our experiments that work with MLP networks, we use a single NVIDIA GeForce RTX 2080 GPU (16 GB) to perform the training and inference. 
For other experiments with larger transformer and convolutional networks---DistilBERT, RoBERTa, and ResNet-18---we use a single NVIDIA A40 GPU (48 GB)
to perform fine-tuning/training and inference.
We use pre-trained checkpoints from prior work whenever possible and limit the amount of training we do ourselves.
In all, we use $\sim$190 GPU hours for our experiments, a breakdown of which is given in Table~\ref{tb:gpu_hours}.

\section{Potential Societal Impacts}
\label{sec:societal_impacts}

Our work in aimed at comparing a given set of models
by shedding light on their potential reliance
on two facets of memorization.
While such evaluation is meant to be used to pick the best model,
(for instance, the least biased one)
one could instead use it to choose the least generalizing model
(the most biased).
This can be especially worrisome in sensitive scenarios
like that of gender bias, where one could easily switch the
order of entropy and MI to obtain a network which
attends to gender information the most.

\section{Dataset Licenses}
\label{sec:dataset_license}

All datasets used in this work are freely publicly available.

\noindent \textbf{Bias-in-Bios:}
MIT License

\noindent \textbf{MNIST:}
GNU General Public License v3.0

\noindent \textbf{IMDb:}
License statement can be found \href{https://help.imdb.com/article/imdb/general-information/can-i-use-imdb-data-in-my-software/G5JTRESSHJBBHTGX#}{here}.

\noindent \textbf{NICO:}
License statement can be found \href{https://nico.thumedialab.com/}{here}.




\end{document}